\documentclass[journal,12pt,onecolumn]{IEEEtran}
\usepackage{algpseudocode}
\usepackage{algorithmicx}
\usepackage{epsfig}
\usepackage{cite}
\usepackage{xcolor}
\usepackage{amsthm}
\usepackage{multirow}
\usepackage{float}
\usepackage{fullpage}
\usepackage[section] {placeins}
\usepackage{amsmath,amsthm,amssymb}

\newcommand\T{\rule{0pt}{2.6ex}}       
\newcommand\B{\rule[-1.2ex]{0pt}{0pt}} 

\newcommand{\algoname}[1]{\textnormal{\textsc{#1}}}
\newcommand{\varA}[1]{{\operatorname{#1}}}
\begin{document}
\newtheorem{theorem}{Theorem}
\title{A New Approach to Dimensionality Reduction for Anomaly
  Detection in Data Traffic}
\author{\IEEEauthorblockN{Tingshan Huang, Harish Sethu and Nagarajan Kandasamy}\\
\thanks{Author Huang is with Akamai Technologies, Cambridge, MA
  02142. Authors Sethu and Kandasamy are with the Department of
  Electrical and Computer Engineering, Drexel University,
  Philadelphia, PA 19104, USA.} 
\thanks{This work was partially supported by the National Science 
  Foundation Award \#1228847.}}
\maketitle

\begin{abstract}
The monitoring and management of high-volume feature-rich traffic in
large networks offers significant challenges in storage, transmission
and computational costs. The predominant approach to reducing these
costs is based on performing a linear mapping of the data to a
low-dimensional subspace such that a certain large
percentage of the variance in the data is preserved in the
low-dimensional representation. This variance-based subspace approach
to dimensionality reduction forces a fixed choice of the number of
dimensions, is not responsive to real-time shifts in observed traffic
patterns, and is vulnerable to normal traffic spoofing. 
Based on theoretical insights proved in this paper, we propose a new
distance-based approach to dimensionality reduction motivated by the
fact that the real-time structural differences between the covariance
matrices of the observed and the normal traffic is more relevant to
anomaly detection than the structure of the training data alone. Our
approach, called the {\em distance-based subspace} method, allows a
different number of reduced dimensions in different time windows and
arrives at only the number of dimensions necessary for effective
anomaly detection. We present centralized and distributed versions of
our algorithm and, using simulation on real traffic traces,
demonstrate the qualitative and quantitative advantages of the
distance-based subspace approach.
\end{abstract}

\thispagestyle{empty}

\section{Introduction}
\label{sec:intro}

Security professionals who monitor communication networks
for malware, errors and intrusions are increasingly dependent on real-time
detection of anomalous behavior in the data traffic \cite{BhuBha2014}. The volume of
data one has to monitor and process for effective real-time
management of networks and systems, however, poses significant Big
Data challenges \cite{YenOpr2013}. Further, many of the anomalies are apparent
only upon an examination of the correlations in the traffic data from multiple
locations. 
State-of-the-art network management in multiple application
scenarios today, therefore, requires the
constant monitoring of a very large number of features at hundreds of
locations across a network and the ability to collect, transmit and
process the data in real-time for near-instantaneous mitigation
actions in case of anomalous patterns indicative of threats. For
example, a data center may use embedded software and hardware sensors to measure various features of the data traffic
across its hundreds of servers (along with other system states such as
temperature, humidity, response time, throughput, processor
utilization, disk I/O, memory, and other network activity) to react
appropriately to anomalous events. Security 
administration of a large network may involve constant monitoring of
its thousands of backbone systems, routers and links to detect a
nascent denial-of-service attack or a developing outage event or a
worm about to propagate itself.

A key step in network management and traffic analysis,
therefore, is {\em dimensionality reduction}, which cuts down the
number of observed features or variables into a smaller set that is
sufficient to capture the information essential to the goals of
network management. 
Principal Component Analysis (PCA) is a widely used tool
that allows us to derive a reduced set of the most significant
uncorrelated features that are linear combinations of the original set of features
\cite{Jolliffe2002}. Given $N$ features, one selects $k \ll N$ most
significant principal components out of $N$ to define a
$k$-dimensional subspace based on observations of normal traffic (it
is generally assumed that training or some reference traffic is
available for purposes of comparing the observed traffic with it.) An underlying
assumption in this approach is that a very small number of principal
components capture most of the variance in the traffic. As a result,
this approach typically chooses $k$ as the number of principal
components which capture a pre-defined percentage (say, 99\%) of
the variance in the normal traffic. Then, a significant deviation in
the projection of the $N$-dimensional observed data onto this
$k$-dimensional reference (normal) subspace can be defined as an
anomaly for purposes of detection \cite{Lakhina2004}. 
In this paper, we refer to this traditional approach to employing PCA
for dimensionality reduction as the {\em variance-based subspace}
method.

There are at least three weaknesses of the traditional variance-based
subspace approach for anomaly detection: (i) the reduced number of
principal components, $k$, is computed based on the structure of
normal traffic when, actually, the structure of the changes between
the observed and the normal traffic is more relevant for choosing the
appropriate number of dimensions for anomaly detection; (ii) a
static determination of $k$ is inadequate at capturing real-time
changes --- the right number of dimensions is often different during
different periods of time depending on the structure of both the
normal and the observed traffic; (iii) the method allows only weak
heuristics because the performance of anomaly detection is very
sensitive to small changes in the number of dimensions chosen for the
normal subspace 
\cite{Ringberg2007}. 
Motivated by the need to address these 
weaknesses, this paper presents a new distance-based approach to
dimensionality reduction for anomaly detection based on a new metric
called the {\em maximum subspace distance}.
In this paper, we refer to
our approach 
as the {\em distance-based subspace} method.  

Normal and anomalous traffic tend to differ in the correlations
between pairs of features of the traffic being monitored. These 
correlations, along with the variance of each monitored feature are
entries in the covariance matrix. This work is additionally motivated by
two observations: (i) anomalies lead to changes in the covariance
matrix of the set of traffic features being monitored, and (ii)
different types of anomalies cause different types of deviations in
the covariance matrix allowing a categorization of the detected
anomaly and an immediate prescription of actions toward threat
mitigation \cite{Yeung2007}. Besides, use of the covariance matrix
requires no assumptions about the distributions of the monitored
features, rendering it a general method for traffic
characterization. 
Rapid detection of structural differences between
two covariance matrices, therefore, is an important goal in modern
anomaly detection and is a key motivation behind the improved method
of dimensionality reduction sought in this work.

The remainder of this paper is organized as
follows. 
Section~\ref{sec:prob} defines the problem statement and
introduces the contributions of this
paper.
Section~\ref{sec:related-work} discusses related work in the  
use of covariance matrices and in subspace methods for anomaly
detection. Section~\ref{sec:metric} describes the assumptions and definitions
underlying the problem and the solution approach used in the
paper. Sections~\ref{sec:getESD} and \ref{sec:getESD-D} describe the
centralized and distributed versions of our algorithms,
respectively. Section~\ref{sec:simulation-results} describes
simulation experiments using real traffic traces which illustrate the
advantages of anomaly detection using our distance-based subspace
method. Finally, Section~\ref{sec:conclusion} concludes the paper.

\section{Problem statement and contributions}
\label{sec:prob}

Let $N$ denote the number of traffic features being monitored in the
network. These $N$ features could all be observations
made at a single location in the network or could be from a
multiplicity of locations where the traffic is being monitored. Assume
that each feature collected or computed from the traffic corresponds to a designated
time slot. We place no assumptions on the selection of traffic
features.

The problem considered in this paper is one of finding an
appropriately reduced set of $k \ll N$ dimensions in real-time, which
may be different during different time intervals depending on the
observed traffic at that time. 
The traditional approach using PCA determines $k$ statically based
only on the training data in order to extract a normal subspace of $k$
dimensions representative of normal traffic, 
even though the best value of $k$ may differ during different periods
of time depending on 
the characteristics of the observed traffic. In contrast, our
distance-based subspace method derives $k$ based on the currently
observed differential between the normal traffic deduced from training
data and the monitored/observed traffic (as opposed to deriving both $k$
and the normal subspace from training data alone). Before any traffic
is observed, one may begin with the choice of a pre-determined
$k$ based on training data as in traditional methods --- but, as soon
as the covariance matrix of observed traffic is computed, our method
allows the subsequent use of only the number of dimensions necessary
at any given time instead of a static determination independent of
current observations.

\subsubsection{Centralized approach} 

In the case in which all $N$ features are observations at a single
node, the covariance matrices can be readily calculated at the node
itself. Also, in the case in which the $N$ features are observations
made at multiple nodes but transmitted and collected at a central
monitoring station, the covariance matrices can be calculated at this
station for analysis. In either of these two cases, we 
can characterize network traffic as a stream of covariance matrices,
one for each designated window of time, and then use the observed
changes in the covariance matrices to infer changes in the system
status, anomalous or otherwise. Let $\Sigma_{A}$ and $\Sigma_{B}$
denote the two $N \times N$ covariance 
matrices that need to be compared to detect changes indicative of an
anomaly. These two matrices could represent real traffic data during
different time windows or one of them could be a reference matrix
representing normal operation without anomalies. The problem becomes
one of devising a measure of the structural difference using a
parsimonious metric that allows efficient computation 
at reasonable accuracy while also serving as a reliable indicator of
anomalies.  

The contribution of this paper begins with the use of a new metric,
which we call the {\em maximum subspace distance}, defined as the largest
possible angle, $\theta_{\mathrm{max}}$, between the subspace composed
of the first $k_A$ principal components of $\Sigma_{A}$ and the
subspace composed of the first $k_B$ principal components of
$\Sigma_{B}$, for all possible values of $k_A$ and $k_B$. The maximum
subspace distance is defined in more detail in
Section~\ref{sec:metric}. 

Finding the subspace (i.e., all the principal components) given a
covariance matrix is computationally expensive since it requires
either singular value decomposition or eigenvalue
decomposition. By avoiding the computation of the entire subspace, by
using only the most significant principal components and by employing
the theoretical insights proved in the Appendix, this
paper contributes a fast approximate algorithm, called
\algoname{getESD}, to estimate a reduced number of principal
components that is sufficiently effective at distinguishing between
the two matrices being compared. This number of principal components
--- which we call the {\em effective subspace   dimension (ESD)} ---
is such that it describes each of the two subspaces enough to gain a
close estimate of the maximum subspace distance. 

\subsubsection{Distributed approach}
\label{subsubsec:dist}

In the case in which the $N$ features are sampled at multiple
locations in the network, the covariance matrix of these features is
not readily computable at any given location. If the feature data
cannot all be transmitted to a central monitoring station due to
communication costs or other reasons typical, we have only the raw features
computed for every time slot based on which we need to 
estimate the maximum subspace distance. Let $M$ denote the number of
consecutive time slots for which these features have to be sampled in
order to ascertain the second-order statistics 
of these features, i.e., the variances and correlations which populate the
covariance matrix of these features. The input to the distributed
algorithm at each node, therefore, is not a stream of covariance
matrices but raw feature data in the form of a stream of vectors, each
of length $M$. 

To address this case, we present a distributed algorithm, called
\algoname{getESD-D}, which avoids the direct computation of the
covariance matrices but nevertheless returns the effective subspace
dimension and an estimate of the maximum subspace distance. The
participating nodes in the \algoname{getESD-D} algorithm deploy the
gossip-based Gaussian mixture learning mechanism to estimate principal
components of the traffic features \cite{vlassis2005gossip}. The
algorithm requires only local computation and some communication
between neighboring nodes. 
The correctness of this algorithm is proved
in the Appendix along with an analysis of its computational and
communication complexity.

Besides offering new theoretical insights into comparisons between
covariance matrices, our work in this paper improves upon the
traditional variance-based approaches to dimensionality reduction in
three ways: (i) our attempt to reduce the number of dimensions is
based on data from both of the two traffic 
streams being compared instead of depending only on some reference
data for a normal traffic stream --- this allows the use of a smaller
and only the
number of dimensions necessary instead of a static pre-determined
number based on the training data alone, (ii) our methods allow a dynamic
real-time computation of the structural changes in network traffic
features and, therefore, allow better characterization and
classification of attack traffic, and (iii) our method makes it
significantly harder for attack traffic to spoof the structure of
normal traffic in order to escape detection. 

\section{Related Work}\label{sec:related-work}

The problem of anomaly detection often invites customized solutions
particular to the type of anomaly that is of interest. For example,
methods reported in \cite{Mandjes2005} and \cite{Freire2008} allow the
detection of load anomalies in voice-over-IP traffic at the network or
the application layer. Similarly, distributed denial-of-service (DDoS)
attacks are the target of detection in 
\cite{Thing2009}, \cite{Xie2009} and \cite{Paschalidis2009}. However,
given that anomalies manifest themselves in multiple and sometimes
uncommon or even unknown ways, a more practical solution to the
problem of anomaly detection is a generic method that can detect
anomalies of all kinds --- common and uncommon, known and unknown.

Developing general anomaly detection tools can be challenging, largely
due to the difficulty of extracting anomalous patterns from huge
volumes of high-dimensional data contaminated with anomalies. Early
anomaly detection methods which used artificial intelligence, machine
learning, or state machine modeling are reviewed in
\cite{Thottan2003}. Examples of later work on developing general
anomaly detection tools include \cite{Lakhina2004,Lakhina2005,Yeung2007,Tavallaee2008,Paschalidis2009,Kind2009,DAlconzo2009,Callegari2011}. There are two broad approaches to anomaly detection that are related
to the work reported in this paper --- those using covariance matrices
directly and those using subspace methods, and occasionally those
which use both these approaches together. 

While the covariance matrix plays a role in many anomaly detection
methods, it was most directly used in \cite{Yeung2007} to detect
flooding attacks based on comparisons between a covariance matrix under
normal conditions (used as the reference matrix) and the observed
covariance matrix. 
The covariance matrix is also used 
directly for anomaly detection in \cite{Tavallaee2008}. 
These methods do not necessarily compete but are complementary to the
approach presented in this paper --- while they address detection, our
paper primarily addresses dimensionality reduction for
detection. Given a reduced number of dimensions, one may choose any of
many detection strategies which depend very strongly on the context
and the application domain. 
Further, as opposed to methods in \cite{Yeung2007} and
\cite{Tavallaee2008} which are based on detecting the changes in
individual entries in the covariance matrix, our method is
additionally able to exploit the underlying correlations
between the changes in the entries to offer a more refined and a more
reliable approach to anomaly detection. 

The variance-based subspace method, based on PCA, was first proposed
for anomaly detection in \cite{Lakhina2004} and later improved in
\cite{Lakhina2005} to explore the deviation in the network-wide
traffic volume and feature distributions caused by anomalies. 
To use this method online in real-time as described in
\cite{Lakhina2004}, one processes each arrival of 
new traffic measurements using the matrix $PP^T$, where $P$ is
composed of the top $k$ principal components representing the normal
traffic pattern. Therefore, for real-time detection using this method,
it is necessary to use a training dataset to determine $P$ before the
detection process along with a certain choice of static $k$ (as
opposed to $k$ being determined dynamically in our distance-based
subspace method.) The scheme proposed in \cite{Lakhina2004} would
separate the high-dimensional space of network traffic into two
subspaces: the normal subspace and the anomalous subspace. The normal
subspace is low-dimensional and captures high variance of normal
traffic data, thus modeling the normal behavior of a network. The
projections of measurement data onto the anomalous subspace are used
to signal, identify and classify anomalies. 

The limitations of the variance-based subspace methods are discussed in
\cite{Ringberg2007}. The simulation results in \cite{Nyalkalkar2011} further
confirm that the effectiveness of the subspace method depends strongly
on the dimension chosen for the normal subspace. In addition,
excessively large anomalies can contaminate the normal subspace and
weaken the performance of the detector. Later work has improved upon
the training process of the subspace method
\cite{Pascoal2012,Mateos2012,Kudo2013}, but choosing the
appropriate dimension for the normal subspace has remained an unmet
challenge. 
The distance-based methods used in the literature also present the
same challenge where the reduced number of dimensions is not adaptive
to real-time data \cite{Bur2010}.
The distance-based subspace method presented in this paper
overcomes this difficulty by allowing the number of dimensions to be
based on both the normal and the observed traffic under examination,
thus adapting constantly to the changing patterns in the observed
traffic to use only the number of dimensions necessary at any given time.

In other related work, PCA-based methods have been decentralized
for a variety of purposes including anomaly detection
\cite{Jelasity2007,Bertrand2012,Meng2012,Huang2007,Wiesel2009}. A
distributed framework for PCA is proposed in \cite{Huang2007} to
achieve accurate detection of network anomalies through monitoring of only
the local data. A distributed implementation of PCA is developed for
decomposable Gaussian graphical models in \cite{Wiesel2009} to allow
decentralized anomaly detection in backbone networks. Distributed
gossip algorithms using only local communication for subspace
estimation have been used in the context of sensor networks
\cite{Boyd2006,Dimakis2010,Li2011}. Our work in this paper
extends the distributed average consensus protocol proposed in
\cite{Li2011} to estimate the principal subspace for the context of
anomaly detection in network traffic data.

\section{The Metric}\label{sec:metric}

Let $N$ denote the number of features in the dataset of interest and
let $\Sigma_{A}$ and $\Sigma_{B}$ denote the two $N \times N$
covariance matrices to be compared. Let $\textbf{a}_1, \dots,
\textbf{a}_N$ and $\textbf{b}_1, \dots, \textbf{b}_N$ denote the
eigenvectors of $\Sigma_{A}$ and $\Sigma_{B}$, respectively. 
In the
following, the operator symbol `$\times$' used between two matrices or
vectors denotes the matrix product. 

\subsection{The subspace distance}
Let $\theta_{k_A, k_B}(A,B)$ denote the angle between the subspace
composed of the first $k_A$ principal components of $\Sigma_{A}$,
$\textbf{a}_1, \dots, \textbf{a}_{k_A}$, and the subspace composed of
the first $k_B$ principal components of $\Sigma_{B}$, $\textbf{b}_1,
\dots, \textbf{b}_{k_B}$. We refer to this angle between the subspaces
as the {\em subspace distance}, which has a range between $0$ to $90$
degrees. We have: 
\begin{equation}
\sin \theta_{k_A, k_B}(A,B) = \,\,\parallel T_{k_A, k_B}(A,B) \parallel
\end{equation}
where $\parallel \cdot \parallel$ is the matrix norm and $T_{k_A,
  k_B}(A,B)$ is the part of $[\textbf{b}_1, \dots, \textbf{b}_{k_B}]$
orthogonal to $[\textbf{a}_1, \dots, \textbf{a}_{k_A}]$. Therefore,
\begin{align}
T_{k_A, k_B}(A,B) &= (I-\sum_{i=1}^{k_A} \textbf{a}_i\times\textbf{a}_i^\prime)[\textbf{b}_1, \dots, \textbf{b}_{k_B}] \\
&= (\sum_{i=k_A+1}^{N} \textbf{a}_i\times\textbf{a}_i^\prime)[\textbf{b}_1, \dots, \textbf{b}_{k_B}].
\end{align}
The sine of the subspace distance captures the norm of $[\textbf{b}_1,
\dots, \textbf{b}_{k_B}]$ not including its projection in
$[\textbf{a}_1, \dots, \textbf{a}_{k_A}]$. The more distinguishable or
orthogonal these two subspaces are to each other, the larger the
subspace distance between them. 

\subsection{The maximum subspace distance}

To compare the two matrices, we quantify the difference between
$\Sigma_{A}$ and $\Sigma_{B}$ as the maximum value of the angle $\theta_{k_A,
  k_B}(A,B)$, where $1\leq k_A \leq N$ and $1\leq k_B \leq N$:
\begin{equation}
\theta_{\mathrm{max}} = \max_{1\leq k_A, k_B \leq N} \theta_{k_A, k_B}(A,B)
\end{equation}
In this paper, we refer to $\theta_{\mathrm{max}}$ as the {\em maximum
  subspace distance}, which serves as our metric for anomaly detection
and quantifies the difference between the two matrices. 

When $k_A$ and $k_B$ hold values that maximize $\theta_{k_A,
  k_B}(A,B)$, the two sets of principal components, $[\textbf{a}_1, \dots, \textbf{a}_{k_A}]$ and $[\textbf{b}_1, \dots,
\textbf{b}_{k_B}]$, can 
be thought of as the distinguishing characteristics of covariance 
matrices $\Sigma_{A}$ and $\Sigma_{B}$. 
Once the maximum subspace distance and the corresponding pair of
dimensions are found, they can be used to characterize the two
datasets with $\Sigma_A$ and $\Sigma_B$ as their covariance matrices,
and distinguish these two datasets from each other. We show in
Section~\ref{sec:simulation-results} how these two sets of characteristics can be
employed for anomaly detection.


Our proposed metric of subspace distance is new, and the closest
related metric in the literature is the \textit{principal
  angle}~\cite{bjorck1973numerical}. 
The subspace distance, as a metric, is different from
principal angle in several ways. Most importantly, the subspace
distance considers the order in importance of each principal component
while the principal angle does not. In addition, unlike in the case of the
subspace distance, the values of the principal angles are not
necessarily dependent on the principal components.


The two sets of principal components used in the definition of the
subspace distance capture the linear correlated pattern of the two
datasets. In addition, the subspace distance depends on the order of
these principal components which represents their order of
importance. As a result, our metric of the maximum subspace distance
serves as a dependable measure of the magnitude of changes in the
pattern between any two datasets.
 
\section{The Centralized Algorithm}\label{sec:getESD}

Given the high computational cost of finding the maximum subspace
distance, in this section, we present an algorithm which estimates the
metric at sufficient accuracy for successful anomaly detection. 
Our solution is based on four key ideas: (i) allowing $k_A = k_B$ in our
search for the maximum subspace distance, $\theta_{\mathrm{max}}$, (ii)
reducing the problem to one of always finding only the first principal
component of a matrix, (iii) using the power iteration
method to approximate the first principal components, and finally, (iv)
using a heuristic to approximately ascertain the value of $\theta_{\mathrm{max}}$.

\subsection{The rationale behind allowing $k_A=k_B$}
In the approach presented in this paper, we limit our search for
$\theta_{\mathrm{max}}$ to only the cases in which $k_A =
k_B$. Our rationale is based on Theorem~\ref{theo:kA-equals-kB}
below. The proof is in the Appendix.

\begin{theorem}
\label{theo:kA-equals-kB}
If for some $k_A$ and $k_B$, $\theta_{k_A,k_B}(A,B) =
\theta_{\mathrm{max}}$, then there exists $k$, $1 \leq k \leq N$,
such that $\theta_{k, k}(A,B) = \theta_{\mathrm{max}}$.
\end{theorem}

Allowing $k_A = k_B$ to find the maximum subspace distance reduces the
search space from $N^2$ to $N$. We refer to the value of $k$ for which 
$\theta_{k, k}(A,B)$ is the maximum subspace distance as the {\em
  optimal subspace dimension}. 

\subsection{Subspace distance and the projection matrix}

We define an $N\times N$ projection matrix $P$ from
$\{\textbf{b}_1,\dots,\textbf{b}_N\}$ to
$\{\textbf{a}_1,\dots,\textbf{a}_N\}$ as: 
\begin{equation}\label{eq:projection}
P_{i,j} = \langle\textbf{a}_i,\textbf{b}_j \rangle = \textbf{a}_i^\prime \textbf{b}_j
\end{equation}
where $\langle \cdot, \cdot \rangle$ represents dot product of two vectors.

According to Theorem 2 below, which we prove in the Appendix, the
smallest singular value of its submatrix $P_{1:k, 1:k}$, consisting of
the first $k$ rows and $k$ columns of $P$, is equal to $\cos(\theta_{k, k}(A,B))$. 

\begin{theorem}
\label{theo:smallest-singular-value}
If an $N\times N$ matrix $P$ is built as in
Eq.~(\ref{eq:projection}), and its submatrix $P_{1:k,1:k}$ has
$\sigma_k(P_{1:k,1:k})$ as its smallest singular value, then
$\sigma_k(P_{1:k,1:k}) = \cos(\theta_{k, k}(A,B))$. 
\end{theorem}

To understand the result in Theorem~\ref{theo:smallest-singular-value}, one can refer to the fact that
the singular values of a projection matrix $P_{1:k,1:k}$ are the
scaling factors of projecting $\textbf{b}_1, \dots, \textbf{b}_k$ onto
$\textbf{a}_1, \dots, \textbf{a}_k$ via matrix $P_{1:k,1:k}$ along
different axes. For example, the largest singular value of
$P_{1:k,1:k}$ shows the largest projection from $\textbf{b}_1, \dots,
\textbf{b}_k$ onto $\textbf{a}_1, \dots, \textbf{a}_k$, which results
in the smallest angle between these two subspaces. The axis in
$\textbf{a}_1, \dots, \textbf{a}_k$ corresponding to the largest
singular value is most parallel to $\textbf{b}_1, \dots,
\textbf{b}_k$. Similarly, the smallest singular value results in an
axis in $\textbf{a}_1, \dots, \textbf{a}_k$ that is most orthogonal to
$\textbf{b}_1, \dots, \textbf{b}_k$, and the resulting angle is the
exact definition of the subspace angle. 

It can be shown, as stated in Theorem~\ref{theo:non-decreasing-to-1} below, that by increasing the
subspace dimension $k$, the largest projection from $\textbf{b}_1,
\dots, \textbf{b}_k$ onto $\textbf{a}_1, \dots, \textbf{a}_k$ is
non-decreasing with $k$ and has 1 as its maximum value. In other
words, the axis in $\textbf{a}_1, \dots, \textbf{a}_k$ becomes more
parallel to $\textbf{b}_1, \dots, \textbf{b}_k$ with increasing $k$.

\begin{theorem}
\label{theo:non-decreasing-to-1}
If an $N\times N$ matrix $P$ is built as in
Eq.~(\ref{eq:projection}), and its submatrix $P_{1:k,1:k}$ has
$\sigma_k(P_{1:k,1:k})$ as its smallest singular value and
$\sigma_1(P_{1:k,1:k})$ as its largest singular value, then:
\begin{align}
&\sigma_1(P_{1:k-1,1:k-1})\leq \sigma_1(P_{1:k,1:k}), ~k=2,\dots,N \label{eq:thm3-1} \\
&\sigma_1(P_{1:k,1:k})\leq 1, ~k=1,\dots,N\label{eq:thm3-2}
\end{align}
\end{theorem}

The proof of Theorem~\ref{theo:non-decreasing-to-1} is in the Appendix. The results in
Eqs.~(\ref{eq:thm3-1}) and (\ref{eq:thm3-2}) indicate that when
$\sigma_1(P_{1:k,1:k})$ is already close enough to 1, further
increasing the number of principal components can only slightly
increase the ability to differentiate between the two subspaces using
the maximum subspace distance as the metric.

\subsection{Estimating the optimal subspace dimension}

Based on our results as stated in Theorems
\ref{theo:kA-equals-kB}--\ref{theo:non-decreasing-to-1}, we develop an 
estimation algorithm for the optimal subspace dimension. It is
straightforward to find the optimal
subspace dimension by obtaining $\textbf{a}_1, \dots, 
\textbf{a}_N$ and $\textbf{b}_1, \dots, \textbf{b}_N$ first, computing
$\theta_{k,k}(A,B)$ for every $k$ from $1$ to $N$ and determining the
$k$ for which $\theta_{k,k}(A,B)$ is the maximum. However, we are
interested in a less computationally expensive method that does not
require full knowledge of $\textbf{a}_1, \dots, \textbf{a}_N$ and
$\textbf{b}_1, \dots, \textbf{b}_N$. Furthermore, our interest is in
searching for a smaller $k$, which we refer to as the {\em effective
  subspace dimension (ESD)}, corresponding to a subspace distance close
enough to the maximum such that this value of $k$ is sufficiently
effective at distinguishing between the two subspaces.

Our algorithm, which we call the \algoname{getESD} algorithm, relies
on the results stated in Theorems~\ref{theo:kA-equals-kB}-\ref{theo:non-decreasing-to-1}. Firstly, because of Theorem~\ref{theo:kA-equals-kB},
we are able to limit our search of the optimal subspace dimension to
cases in which $k_A = k_B = k$. Secondly, since the values of subspace
distance, $\theta_{k,k}(A,B)$, depend on singular values of submatrix
$P_{1:k,1:k}$, as stated in Theorem~\ref{theo:smallest-singular-value}, we can compute the subspace
distance $\theta_{k,k}(A,B)$ with only the knowledge of the first $k$
eigenvectors of $\Sigma_A$ and $\Sigma_B$. In other words, if we have
already tried values of $k$ from $1$ to $K$, the pair of ($K+1$)-th
eigenvectors is all the additional information we need to get
$\theta_{K+1,K+1}(A,B)$. 

Finally, because of the property of singular values of submatrix
$P_{1:k,1:k}$, as stated in Theorem~\ref{theo:non-decreasing-to-1}, $\sigma_1(P_{1:k,1:k})$ is
non-decreasing to $1$. The closer $\sigma_1(P_{1:k,1:k})$ is to $1$,
the less distinguishable these two subspaces become. As a result, we
set a threshold on $\sigma_1(P_{1:k,1:k})$, and stop the algorithm
when $\sigma_1(P_{1:k,1:k})$ is above the threshold $1-\epsilon$. Of
course, a higher threshold leads to a wider search range and a closer
approximation of the optimal subspace dimension. 

Overall, our algorithm computes an approximate value of $\theta_{k,k}(A,B)$
beginning with $k = 1$ and increases $k$ at each step; the algorithm stops when the subspace distance,
$\theta_{k,k}(A,B)$, has dropped and the largest singular value of $P_{1:k,1:k}$ is in a pre-defined neighborhood of 1.
This observed $\theta_{k,k}(A,B)$ is used as the {\em
  estimated} maximum subspace distance and the corresponding $k$
becomes the effective subspace dimension. 

Fig.~\ref{fig:eff-alg} presents the pseudo-code of our algorithm,
\algoname{getESD}, which returns ESD, the effective subspace
dimension, and $\theta_{\text{max}}$, the estimated maximum subspace
distance, given two covariance matrices, $\Sigma_A$ and $\Sigma_B$, and
$\epsilon$, which defines the threshold at which the algorithm
stops. Note that the $k$-th eigenvectors of
$\Sigma_A$ and $\Sigma_B$ are not computed until we are in the $k$-th
iteration of the {\tt while} loop. Our implementation of this
algorithm uses the power iteration method \cite{golub2012matrix}. For
each iteration in our algorithm, we add one row and one column to the
projection matrix along with the updated pair of eigenvectors. During the
$k$-th iteration of the {\tt while} loop, we compute $\textbf{a}_k$
and $\textbf{b}_k$, i.e., the $k$-th eigenvectors of $\Sigma_A$ and
$\Sigma_B$, using the power iteration method. We then construct
$P_{1:k,1:k}$ by adding row vector
$[\textbf{a}^{\prime}_k\textbf{b}_{1}, \dots,
\textbf{a}^{\prime}_k\textbf{b}_{k-1}]$ to the bottom of
$P_{1:k-1,1:k-1}$, and column vector
[$\textbf{a}^{\prime}_1\textbf{b}^{\prime}_{k}, \dots,
\textbf{a}^{\prime}_k\textbf{b}^{\prime}_{k}$] to the right. We then
use the power iteration method again to calculate the singular values
$\sigma_1(P_{1:k,1:k})$ and $\sigma_k(P_{1:k,1:k})$. The algorithm is
run until the aforementioned stopping criteria is met.

The \algoname{getESD} algorithm achieves a significantly higher
accuracy over our previous work, a heuristic \cite{Huang2015}. As reported in \cite{Huang2015}, the heuristic is
able to estimate the subspace distance with a percentage error
which frequently reaches above 1\% and sometimes close to 5\%. For the same
data, however, the percentage error using the \algoname{getESD}
algorithm never exceeds 0.051\%, a significant improvement.

\begin{figure}
\begin{algorithmic}
\Procedure{getESD}{$\Sigma_{A}$, $\Sigma_{B}$, $\epsilon$}
	\State $k\gets 1$\Comment{number of Principal Components (PCs)}
	\State $\theta\gets 0$\Comment{angle between two subspaces}
	\State $\hat{\Sigma}_A\gets \Sigma_A$\Comment{projection of $\Sigma_A$ on its last $N-k$ PCs}
	\State $\hat{\Sigma}_B\gets \Sigma_B$\Comment{projection of $\Sigma_B$ on its last $N-k$ PCs}
	\State $\theta_{\mathrm{max}} \gets 0$\Comment{maximum angle observed}
	\State {\em ESD} $ \gets 0$\Comment{value of $k$ corresponding to $\theta_{\mathrm{max}}$}
	\While{($k\le N$)}
		\State $\textbf{a}_k\gets $ estimated first PC of $\hat{\Sigma}_A$\Comment{k-th PC of $\Sigma_A$}
		\State $\textbf{b}_k\gets $ estimated first PC of $\hat{\Sigma}_B$		\Comment{k-th PC of $\Sigma_B$}
		\State $\text{construct}~P^{\prime}_{1:k,1:k}P_{1:k,1:k}$	
		\State $\sigma_1(P_{1:k,1:k}) \gets \sqrt{\lambda_1(P^{\prime}_{1:k,1:k}P_{1:k,1:k})}$
		\State $\sigma_k(P_{1:k,1:k}) \gets \sqrt{\lambda_k(P^{\prime}_{1:k,1:k}P_{1:k,1:k})}$
		\State $\theta^{\prime} \gets \arccos(\sigma_k(P_{1:k,1:k}))$
		\If{$(\theta^{\prime} < \theta~\&~\sigma_1(P_{1:k,1:k})>1-\epsilon)$}
			\State \Return ({\em ESD}, $\theta_{\mathrm{max}}$)
		\EndIf
		\If{$(\theta > \theta_{\mathrm{max}})$}
			\State $\theta_{\mathrm{max}} \gets \theta$
			\State {\em ESD} $ \gets k$
		\EndIf
		\State $\hat{\Sigma}_A\gets \hat{\Sigma}_A-\textbf{a}_k\times(\textbf{a}_k^\prime\times \hat{\Sigma}_A)$
		\State $\hat{\Sigma}_B\gets \Sigma_B-\textbf{b}_k\times(\textbf{b}_k^\prime\times \hat{\Sigma}_B)$
		\State $k\gets k+1$			
		\State $\theta \gets $ $\theta^{\prime}$
	\EndWhile
	\State \Return ({\em ESD}, $\theta_{\mathrm{max}}$)
\EndProcedure
\end{algorithmic}
\caption{The \algoname{getESD} algorithm: An efficient algorithm for
  finding an effective subspace dimension (ESD) and the corresponding
  estimate of the maximum subspace distance between the two given
  matrices, $\Sigma_{A}$ and $\Sigma_{B}$.}\label{fig:eff-alg} 
\end{figure}

\subsection{Complexity analysis}

By the naive approach, we would first calculate the two sets of
eigenvectors and then the subspace distance for every possible pair of
$(k_A, k_B)$ to ascertain the maximum subspace distance. The
complexity of calculating the two sets of eigenvectors is
$\mathcal{O}(N^3)$. The complexity of calculating the subspace
distance for every possible pair of $(k_A, k_B)$ is
$N^2*\mathcal{O}(N^3)=\mathcal{O}(N^5)$. The computational complexity
of the naive method which computes all eigenvectors to find the
optimal subspace dimension, therefore, is $\mathcal{O}(N^5)$.
 
Theorem~\ref{theo:getESD-complexity} states the computational
complexity of the \algoname{getESD} algorithm.

\begin{theorem}
\label{theo:getESD-complexity}
The \algoname{getESD} algorithm in Fig. \ref{fig:eff-alg} for $N$
nodes has a computational  complexity of $\mathcal{O}(Zk^3+k^2N)$,
where $k$ is the resulting effective subspace dimension and $Z$ is the
upper bound on the number of iterations in the power method. 
\end{theorem}

We prove Theorem~\ref{theo:getESD-complexity} in the Appendix. This
shows that the \algoname{getESD} algorithm is significantly more
computationally efficient compared with the naive method which
computes the exact optimal subspace dimension. 
If a variance-based approach were to be used in each time window to
dynamically compute the number of dimensions, the time complexity
(determined by having to compute the full eigenvector) would be
$\mathcal{O}(N^3)$, significantly higher than that for the
\algoname{getESD} algorithm.
We will demonstrate in Section~\ref{sec:simulation-results} that the
approximate result computed by the \algoname{getESD} algorithm suffices to detect anomalies.

\section{Distributed Algorithm}\label{sec:getESD-D}

When a multiplicity of points in a network are being monitored for
anomalies, the data required for the computation of covariance
matrices has to be collected at multiple nodes, some of them
geographically distant from each other. The centralized algorithm in
Section~\ref{sec:getESD} would require a monitoring station to collect
all the data from the collection nodes, compute the covariance
matrices, and then process the stream of matrices for anomalies. The
prohibitive logging and communication costs in this case form the
motivation behind our distributed algorithm to allow an estimation
of the maximum subspace distance without a central station. 

In the distributed case, no node can possess the knowledge of the
entire covariance matrix (which requires all of the data from all of
the nodes) but each node can update and maintain its corresponding
entry in the eigenvector. As a result, each node will need to
collaborate with its neighbors to perform eigenvector/eigenvalue
estimation in a distributed fashion.

\subsection{Assumptions and system model}

\begin{figure}[!t]
\begin{algorithmic}
\Procedure{AverageConsensus}{$x_n$,$W$}
	\State $m_n^{(0)} \gets x_n$\Comment{initialize estimate}	
	\State $k \gets 0$\Comment{step}
	\For{$k=1:S$}
		\State sends $m_n^{(k)}$ to its neighbors
		\State receives $m_i^{(k)}$ from its neighbors, $i \in \text{Neighbor}(n)$
		\State $m_n^{(k+1)} \gets \sum_{i \in \text{Neighbor}(n)}	 W_{n,i}m_i^{(k)}	$
	\EndFor
	\State \Return $m_n^{(k)}$
\EndProcedure
\end{algorithmic}
\caption{The average consensus procedure.}\label{fig:alg_average_consensus} 
\end{figure}
 
Let $N$ be the total number of features being monitored. In practice,
different nodes will have different numbers of features to
monitor. However, for purposes of simplicity and clarity, we will illustrate our distributed algorithm assuming there is
one feature monitored at each node --- so, in this section, we will
also assume that there are $N$ nodes. As described in
Section~\ref{subsubsec:dist}, let $M$ denote the number of
consecutive time slots for which each of these $N$ features is sampled
to compute a covariance matrix (any two features have to be
observed for a certain length of time to ascertain the correlation
between them). The input to the distributed algorithm, therefore, is a
stream of vectors of length $M$ each, with each entry being a feature
sampled at a certain time slot.  

Collectively, these vectors form an $N \times M$ matrix, $X$, with
data collection of the $n$-th feature $\textbf{x}_n$ available only at node
$n$ (here, $\textbf{x}_n$ is the $n$-th row of raw data). For
an eigenvector $\textbf{v}$ of $X$, ${v}_n$ is estimated and stored at
the $n$-th node.  

We also assume a weighted connection matrix $W$ associated with an
undirected graph composed of these $N$ nodes. Besides a self-loop for
each node, an edge exists between any two nodes as long as they can 
communicate with each other directly (we make no assumptions on the
protocol layer for this communication, although the detection
algorithm will achieve better performance at the IP layer than at the
application layer). Each edge is associated with a weight, and the
total weight of all edges connected to the same node is one. This
makes the column sums and row sums of $W$ equal to one.

We also make use of a distributed routine to compute the average of
certain values at all nodes but one which runs only locally at each
node. This routine, a straightforward extension of the distributed average
consensus algorithm in \cite{Li2011} to accommodate for the
constraints placed on row sums and column sums of $W$, is shown in
Fig.~\ref{fig:alg_average_consensus}.  If each of the $N$ nodes holds
the value $x_n$, $n=1,\dots,N$, this routine computes the average $m =
\frac{1}{N}\sum_{n=1}^N x_n$ at each node locally.

%
%
%
%

\begin{figure}[!t]
\begin{algorithmic}
\Procedure{PowerIteration}{$C$, $\epsilon$}
	\State $\textbf{v}^{(0)} \gets$ randomly chosen $ N\times 1$ vector \Comment{initialize}	
	\State $k \gets 0$\Comment{step}
	\While{True}
		\State $\textbf{v}^{(k+1)}\gets C\textbf{v}^{(k)}$		
		\If{$\parallel\textbf{v}^{(k+1)}-\textbf{v}^{(k)} \parallel < \epsilon$}
			\State \Return $\textbf{v}^{(k+1)}/\parallel \textbf{v}^{(k+1)} \parallel$
		\EndIf
		\State $k\gets k+1$			
	\EndWhile
\EndProcedure
\end{algorithmic}
\caption{The centralized power iteration procedure.}\label{fig:power-itr} 
\end{figure}

\subsection{Distributed power iteration method}\label{subsec:distr-power}

The goal of a distributed power iteration method is to allow nodes
that are remotely located to cooperate in their estimation of the most
significant principal component of a dataset. Fig.~\ref{fig:power-itr}
presents the pseudo-code for a centralized power iteration method
which estimates the first principal component $\textbf{v}$ given a
covariance matrix $C$. The method rests on the fact that the first
eigenvalue of a given matrix $C$ can be estimated by
$\lim_{k\to\infty} \frac{C^k \textbf{v}}{\parallel C^k
  \textbf{v}\parallel}$, where $\textbf{v}$ is a random vector. We
refer readers to \cite{golub2012matrix} for more details about the
power iteration method. 

%

In the following, we develop and describe a distributed version of the
power iteration method. First, we describe how power iteration can be
performed without involving the covariance matrix $C$. In the $k$-th
step of the centralized power iteration method shown in
Fig.~\ref{fig:power-itr}, $\textbf{v}^{(k+1)}$ needs to be updated by
multiplying with $C$. In the absence of $C$, $\textbf{v}^{(k+1)}$ can
be computed as  
\begin{align*}
\textbf{v}^{(k+1)} &= C\textbf{v}^{(k)}
= \frac{1}{M-1}XX^{\prime}\textbf{v}^{(k)}\\
&= \frac{1}{M-1}X[\textbf{x}_1^{\prime} \dots \textbf{x}_N^{\prime} ]\textbf{v}^{(k)}
= \frac{1}{M-1}X\sum_{i=1}^N \textbf{x}_i^{\prime}  {v}^{(k)}_i
\end{align*}
The $n$-th entry of the
estimated principal component at $(k+1)$-th step is:
\begin{align}\label{eq:est_PC_n}
{v}_n^{(k+1)} = \frac{1}{M-1} \textbf{x}_n \textbf{z}^{(k)}
\end{align}
where $\textbf{z}^{(k)}=\sum_{n=1}^N \textbf{x}_n^{\prime}
{v}^{(k)}_n$. Since the $n$-th node has access to $\textbf{x}_n$ and
$\textbf{x}_n^{\prime} {v}^{(k)}_n$, it is able to compute
${v}_n^{(k+1)}$ after it uses the average consensus process to get
$\textbf{z}^{(k)}$.  
The norm of ${v}^{(k)}$ and ${v}^{(k+1)} - {v}^{(k)}$ can be achieved
by applying average consensus on $({v}_n^{(k)})^2$ and
$({v}_n^{(k+1)}-{v}_n^{(k)})^2$. 

Our distributed version of the power iteration method is shown in
Fig.~\ref{fig:dist-power-itr}. Note that the input to this algorithm
is $\textbf{x}_n$, raw data visible to the node running the algorithm,
with the method requiring no knowledge of covariance matrix
$C$. Besides $\textbf{x}_n$, the algorithm takes an additional input,
$\hat{\textbf{x}}_n$, which we explain in
Section~\ref{subsec:getESD-D}. For now, it suffices to know that
$\hat{\textbf{x}}_n=\textbf{x}_n$ for the estimation of ${v}_n$. The
estimated principal component is normalized before it is returned, and
each node only needs to estimate one entry of $\textbf{v}^{(k)}$ in
each step.

\begin{figure}
\begin{algorithmic}
\Procedure{PowerIteration-D}{$\textbf{x}_n$, $\hat{\textbf{x}}_n$, $\epsilon$}
	\State ${v}_n^{(0)}\gets$ randomly chosen value \Comment{initialize}	
	\State $k \gets 0$\Comment{step}
	\While{True}
		\State $\textbf{z}^{(k)} \gets N \times \text{AverageConsensus}(\textbf{x}_n^{\prime} {v}_n^{(k)},W) $
		\State ${v}_n^{(k+1)} \gets \frac{1}{M-1} \hat{\textbf{x}}_n \textbf{z}^{(k)}$ 
		
		\State $e_n^{(k)} \gets \text{AverageConsensus}((v_n^{(k+1)}-v_n^{(k)})^2,W)$ 
		\If{$e_n^{(k)}<\epsilon$}
			\State $l_n \gets \text{AverageConsensus}((v_n^{(k+1)})^2,W)$
			\State \Return $\frac{1}{\sqrt{l_n}}{v}_n^{(k+1)}$
		\EndIf
		\State $k\gets k+1$			
	\EndWhile
\EndProcedure
\end{algorithmic}
\caption{The distributed power iteration method.}\label{fig:dist-power-itr} 
\end{figure}

\subsection{Evaluation of convergence and accuracy}\label{sec:eval-dist}

In this subsection, we briefly report on the convergence of the
distributed power iteration method and its accuracy. We use real
data traffic traces --- anonymized passive traffic
traces from one of CAIDA's monitors on high-speed Internet backbone
links, equinix-chicago-dirA \cite{CAIDA}. 
Use of real traffic traces allows an examination of the algorithms in
real contexts where there is always some degree of both noise and
unmarked anomalies present in the data.

We consider the histogram of packet sizes and protocol of the packets
flowing over a single link within an interval of 25ms. We use 75 bins
for packet sizes and 4 for protocol types, leading to a feature vector
of length 79. For the trace lasting 1 minute, we obtain around 2400
examples of the feature vectors and use them to calculate a covariance
matrix of these 79 features.

For lack of publicly available traffic traces paired with a large
network graph, we now lay this dataset on a simulated network graph
composed of 79 nodes, generated using the scale-free
Barab\'asi--Albert (BA) model where, after initialization, each new 
node added to the network is attached to 2 existing nodes. To clearly
isolate the performance of the distributed power iteration method, we
assume a simple scenario with each node responsible for the collection
of one type of feature and the computation of one histogram. The
weights for edges, including self loops, are randomly chosen to satisfy
the requirement that the weight matrix has unit row sums and unit
column sums.  

Since the family of average consensus algorithms is already
well-studied \cite{Li2011}, we focus here on the distributed power
iteration method assuming that the distributed average consensus
method generates an answer within an error margin of 5\%. 

\begin{figure}[!t]
\centering
\epsfig{file=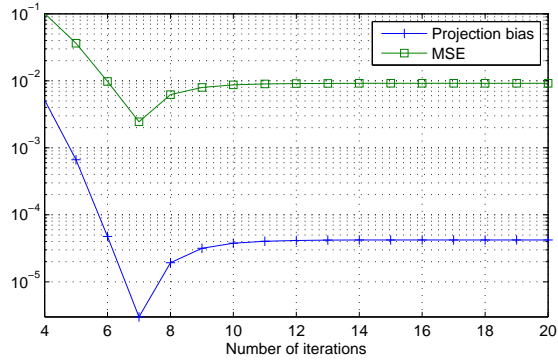,width=3.4in}
\caption{Error of the estimated principal component using distributed
  power iteration using different numbers of iterations in the
  distributed power iteration method. The errors are reported in terms
  of two metrics: the projection bias and the mean squared error
  (MSE). Note that the error after convergence (at iteration 12 and
  beyond) is the same as the error with the centralized version.}
\label{fig:result_dPower_converge_exponential}
\end{figure}

We use two metrics to evaluate the estimation result of our algorithm for distributed power iteration. 
The first, \textit{projection bias}, quantifies the bias of the
estimated principal component in the direction of the actual principal
component. The second metric, mean squared error (MSE), quantifies the
closeness of the estimate to the actual principal
component. Fig.~\ref{fig:result_dPower_converge_exponential} shows 
these metrics for the distributed power iteration method using
simulation results for varying numbers of iterations.

Since the actual principal component is not available to our
algorithm, it does not know to stop when both the projection bias and MSE are the lowest (as at
iteration 7 in Fig.~\ref{fig:result_dPower_converge_exponential}), but
continues on to converge to some value other than the better answer it had 
already found. However, the algorithm does converge at an estimate
which is the same as that estimated by the centralized version when
all of the data is available to all the nodes (in our example in
Fig.~\ref{fig:result_dPower_converge_exponential}, this happens after
iteration 12.) Since the centralized version of our algorithm also
stops at an approximation, neither of the two versions arrives at
exactly the correct answer but they both achieve a very low error in
their estimates. Note that 
Fig.~\ref{fig:result_dPower_converge_exponential} also offers a
comparison in the accuracy achieved by the centralized and the
distributed versions, since the estimation of the principal component
is what differentiates the \algoname{getESD} and \algoname{getESD-D}
algorithms. The accuracy of the distributed \algoname{getESD-D}
algorithm given
enough iterations to allow convergence of distributed power iteration
is the same as the accuracy achieved by the centralized
\algoname{getESD} algorithm.

While the number of required iterations depends
on the threshold for the stopping condition, we find that, on average,
the result converges after around 10 distributed power iterations and
that an equivalent or better result is achieved with as few as 6
iterations. Fig.~\ref{fig:result_dPower_converge_exponential} also
shows that the estimated principal component is almost in the same direction
as the actual principal component with a mean squared error below
$1\%$. Finally, it is worth noting that by exploiting results from
iterations in previous time windows, one can effectively reduce the
number of iterations per time window to an average of $1$.

\subsection{The \algoname{getESD-D} algorithm}
\label{subsec:getESD-D}

In this section, we present the \algoname{getESD-D} algorithm, the
decentralized implementation of our algorithm in
Fig.~\ref{fig:eff-alg}. In each iteration of the {\tt while} loop, the
first principal component and the corresponding eigenvalue are
estimated by the distributed power iteration method.  

After obtaining $\textbf{a}_k$ and $\textbf{b}_k$, $P_{1:k,1:k}$ can
be constructed as follows. Since $P_{i,j}=\sum_{n=1}^N
\textbf{a}_i(n)\textbf{b}_j(n)$, entries of row vector
$[\textbf{a}^{\prime}_k\textbf{b}_{1}, \dots,
\textbf{a}^{\prime}_k\textbf{b}_{k-1}]$ and column vector
$[\textbf{a}^{\prime}_1\textbf{b}_{k}, \dots,
\textbf{a}^{\prime}_k\textbf{b}_{k}]^{\prime}$ can be calculated by
calling the distributed algorithm for average consensus on
$\textbf{a}_i(n)\textbf{b}_j(n)$. Then we can add the row vector
$[\textbf{a}^{\prime}_k\textbf{b}_{1}, \dots,
\textbf{a}^{\prime}_k\textbf{b}_{k-1}]$ to the bottom of
$P_{1:k-1,1:k-1}$, and column vector $[\textbf{a}^{\prime}_1\textbf{b}_{k}, \dots,
\textbf{a}^{\prime}_k\textbf{b}_{k}]^{\prime}$ to its right to construct $P_{1:k,1:k}$.

Note that at the end of each {\tt while} loop of Fig.~\ref{fig:eff-alg}, the
part corresponding to $\textbf{a}_k$ and $\textbf{b}_k$ are subtracted
from covariance matrices $\hat{\Sigma}_A$ and
$\hat{\Sigma}_B$. Without the subtraction, the eigenvector
$\textbf{v}^{(k+1)}$ can be equivalently estimated using distributed
power iteration as follows. Let $\textbf{a}_k$ be the
estimate for the $k$-th eigenvector of dataset $X$, then 
\begin{align*}
\textbf{v}^{(k+1)} &= (C - \textbf{a}_k\textbf{a}_k^{\prime}C)\textbf{v}^{(k)}\\
&= C\textbf{v}^{(k)} - \frac{1}{M-1}\textbf{a}_k\textbf{a}_k^{\prime}XX^{\prime}\textbf{v}^{(k)}\\
&= C\textbf{v}^{(k)} - \frac{1}{M-1}\textbf{a}_k   \left(\sum_{i=1}^N \textbf{a}_k(i)\textbf{x}_i\right)   \left(\sum_{j=1}^N \textbf{x}_n^{\prime} {v}_n^{(k)} \right)\\
&= C\textbf{v}^{(k)} - \frac{1}{M-1}\textbf{a}_k   \left(\sum_{i=1}^N \textbf{a}_k(i)\textbf{x}_i\right)  \textbf{z}^{(k)},
\end{align*}
where $\textbf{z}^{(k)}$ is defined as in Eq.~(\ref{eq:est_PC_n}). 
In particular, the $n$-th entry of the eigenvector $\textbf{v}^{(k+1)}$ can be estimated by:
\begin{align*}
{v}_n^{(k+1)}  &=  \left(C\textbf{v}^{(k)}\right)_n- \frac{1}{M-1}\textbf{a}_k(n)   \left(\sum_{i=1}^N \textbf{a}_k(i)\textbf{x}_i\right)  \textbf{z}^{(k)}\\
&= \frac{1}{M-1} \textbf{x}_n\textbf{z}^{(k)}- \frac{1}{M-1}\textbf{a}_k(n)   \left(\sum_{i=1}^N \textbf{a}_k(i)\textbf{x}_i\right)  \textbf{z}^{(k)}\\
&= \frac{1}{M-1} (\textbf{x}_n - \textbf{a}_k(n) f(\textbf{a}_k,X))\textbf{z}^{(k)}
\end{align*}
where $f(\textbf{a}_k,X) = \sum_{i=1}^N \textbf{a}_k(i)\textbf{x}_i$. 
As a result, a copy of the readings at node $n$, denoted by
$\hat{\textbf{x}}_n$, can be updated using $\textbf{x}_n -
\textbf{a}_k(n) f(\textbf{a}_k,X)$ to allow equivalent estimation as
in Fig.~\ref{fig:eff-alg}. 

The distributed algorithm that combines the aforementioned computations is shown in Fig.~\ref{fig:distr-eff-alg}.

\begin{figure}
\begin{algorithmic}
\Procedure{getESD-D}{$\textbf{x}_n$, $\textbf{y}_n$, $\epsilon$}
	\State $k\gets 1$\Comment{number of Principal Components (PCs)}
	\State $\theta\gets 0$\Comment{angle between two subspaces}
	\State $\theta_{\mathrm{max}} \gets 0$\Comment{maximum angle observed}
	\State $\hat{\textbf{x}}_n \gets \textbf{x}_n$ \Comment{copy of $\textbf{x}_n$}
	\State $\hat{\textbf{y}}_n \gets \textbf{y}_n$ \Comment{copy of $\textbf{y}_n$}
	\State {\em ESD} $ \gets 0$\Comment{value of $k$ corresponding to $\theta_{\mathrm{max}}$}
	\While{($k\le N$)}
		\State ${a}_k(n)\gets $ \algoname{PowerIteration-D}($\textbf{x}_n$, $\hat{\textbf{x}}_n$, $\epsilon$)
		\State ${b}_k(n)\gets $ \algoname{PowerIteration-D}($\textbf{y}_n$, $\hat{\textbf{y}}_n$, $\epsilon$)
		\State $\text{construct}~P^{\prime}_{1:k,1:k}P_{1:k,1:k}$	
		\State $\sigma_1(P_{1:k,1:k}) \gets \sqrt{\lambda_1(P^{\prime}_{1:k,1:k}P_{1:k,1:k})}$
		\State $\sigma_k(P_{1:k,1:k}) \gets \sqrt{\lambda_k(P^{\prime}_{1:k,1:k}P_{1:k,1:k})}$
		\State $\theta^{\prime} \gets \arccos(\sigma_k(P_{1:k,1:k}))$
		\If{$(\theta^{\prime} < \theta~\&~\sigma_1(P_{1:k,1:k})>1-\epsilon)$}
			\State \Return ({\em ESD}, $\theta_{\mathrm{max}}$)
		\EndIf
		\If{$(\theta > \theta_{\mathrm{max}})$}
			\State $\theta_{\mathrm{max}} \gets \theta$
			\State {\em ESD} $ \gets k$
		\EndIf
		\State $f_X\gets \algoname{AverageConsensus}(\textbf{a}_k(n)\textbf{x}_n,W) $
		\State $\hat{\textbf{x}}_n\gets \hat{\textbf{x}}_n-\textbf{a}_k(n)f_X$
		\State $f_Y \gets \algoname{AverageConsensus}(\textbf{b}_k(n)\textbf{y}_n,W) $
		\State $\hat{\textbf{y}}_n\gets \hat{\textbf{y}}_n-\textbf{b}_k(n)f_Y$
		\State $k\gets k+1$			
		\State $\theta \gets $ $\theta^{\prime}$
	\EndWhile
	\State \Return ({\em ESD}, $\theta_{\mathrm{max}}$)
\EndProcedure
\end{algorithmic}
\caption{The distributed algorithm, \algoname{getESD-D} running at node $n$ to
  find an effective subspace dimension (ESD) and the corresponding
  estimate of the maximum subspace distance between the two given
  datasets, $X$ and $Y$.}\label{fig:distr-eff-alg} 
\end{figure}
\subsection{Complexity analysis}
Theorem~\ref{theo:getESD-D-comp-complexity} below, proved in the
Appendix, addresses the computational complexity the \algoname{getESD-D} algorithm.

\begin{theorem}
\label{theo:getESD-D-comp-complexity}
The computational complexity of \algoname{getESD-D} 
is $\mathcal{O}(kpM\Delta S)$ for $M$ measurements at each node,
where $S$ and $p$ are upper bounds on the number of steps it takes for
the convergence of average consensus and the \algoname{getESD-D}
methods, respectively.
\end{theorem}

Note that in a central setting, the computational cost for
calculating covariance matrix is $\mathcal{O}(N^2M)$, and is
$\mathcal{O}(ZN^2)$ for $Z$ steps of the power method. The messages
with information about principal components to all nodes would be
$\mathcal{O}(N^3)$ in the best case and $\mathcal{O}(N^4)$ in the worst case.  

Theorem~\ref{theo:getESD-D-comm-complexity} below, also proved in the
Appendix, addresses the communication complexity the \algoname{getESD-D} algorithm.

\begin{theorem}
\label{theo:getESD-D-comm-complexity}
The communication complexity of \algoname{getESD-D}  is 
$\mathcal{O}(kM\Delta + kp\Delta S)$ for $M$ measurements at each node
in a network with a maximum degree of $\Delta$, where $S$ and $p$ are
upper bounds on the number of steps it takes for the convergence of
average consensus and the \algoname{getESD-D} methods, respectively. 
\end{theorem}

Note that the number of steps for the average
consensus process, $S$, depends on the ratio of
$\lambda_2(W)/\lambda_1(W)$, the second and first eigenvalues of $W$:
a smaller ratio results in a smaller $S$. For a sparse graph, such a
condition can be easily met. Similarly, the upper 
bound on the number of iterations in the distributed power method,
$p$, relies on the ratio of $\lambda_2(C)/\lambda_1(C)$, the second and
first eigenvalues of $C$. In our tests with CAIDA traffic data \cite{CAIDA}, the ratio is as
small as $3\varA{E--}4$, which means that only a couple of iterations
for a 7000-dimensional dataset are enough to reach convergence.

\section{Effectiveness of Anomaly Detection}\label{sec:simulation-results}

Assume the dimension of data is $N$. Using training data, a normal
subspace $U_k$ is then constructed with the top $k \ll N$ principal
components. Since most of the normal traffic patterns lie within the
normal subspace $U_k$, projections of data with no anomalies onto the
normal subspace should retain most of the data. However, projections
of anomalous data, whose characteristics do not fall entirely within
the aforementioned normal subspace, would retain only a portion of the
observed data. Therefore, for purposes of anomaly detection using the
variance-based subspace method, the {\em projection residual} of any
given length-$N$ vector $\textbf{x}$, denoted by $r(\textbf{x})$, is
calculated as follows to yield a metric of the anomaly:
\begin{align*}
r(\textbf{x}) = \, \parallel (I-U_k U_k^T) \textbf{x}  \parallel ,
\end{align*}
where $I$ is the identity matrix. The detection process simply
compares the projection residual with a pre-defined threshold and
triggers an alarm if it is above the threshold \cite{Lakhina2005}.
 
The projection residual is the residual of the data after it has been
projected onto the normal subspace: if the data is normal, the
projection residual should be small relative to the case when the data
consists of anomalies. In this paper, we will use the projection
residual as a measure of the degree of anomalousness in the
traffic. Note that in the distributed scenario, the projection
residual can be computed at each node by calling the average consensus
method on $U_k(n,:)x_n$.


\subsection{Anomaly detection using the projection residual}\label{sec:results}

To test our distance-based subspace approach for anomaly detection, we
need labeled traffic data with connections that are already identified as
either normal or anomalous. We use the Kyoto2006+ dataset,
which includes 24 features of raw traffic data obtained by honeypot
systems deployed at Kyoto University \cite{KyotoData}. Traffic
features include, among others, source/destination IP address and port
numbers of a connection, its duration, and bytes that are
received/sent. Four additional features are included for each
connection, indicating whether the session is part of an attack or
not. In our simulation, we use these 4 features for verification and
the other 20 features for data analysis. 

Since raw features often show low correlation for useful anomaly
detection, we extract the entropy of each traffic feature within
non-overlapping time windows and use the length-20 entropy vector for
each time window as the new traffic feature. In the dataset, there are
about 100 new connections every minute, with most connections lasting
around 1 minute. As a result, we use a time window lasting 5 minutes
and extract the entropy of around 500 connections in each time
window. 
The choice of a window size for optimal effectiveness depends not only
on the traffic intensity but also on the features being monitored ---
extracting some features may require more traffic history than certain
others. However, a larger window size need not increase the delay in
anomaly detection, except upon start-up. To detect anomalies as
rapidly as possible, one can use our algorithms with sliding
(overlapping) time windows to update the sample covariance matrix and
run the detection every time the window moves.

In subspace methods, we use training data to construct a subspace
consisting of its first few principal components, which is assumed to
be the normal subspace. We then project the observed/test data, which
may contain varying numbers of anomalous connections at any given
time, onto this normal subspace. This yields the projection 
residual, our metric of anomalousness. In our experiments using the
labeled Kyoto data, we generate the training data using only the
connections that are not labeled as anomalous. The test data, on the
other hand, contains a varying number of anomalous connections.

\begin{figure}[!t]
\centering
\epsfig{file=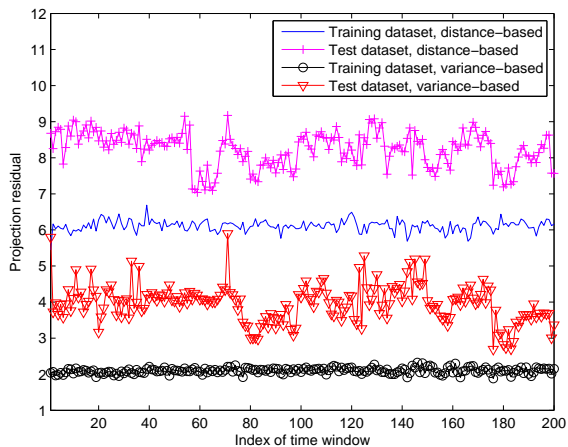,width=3.4in}
\caption{A comparison of projection residuals when the normal subspace
  dimension is determined by the \algoname{getESD} algorithm
  (distance-based) and by the variance-based subspace method
  (variance-based). The dimension used in the variance-based method is
  the number required to capture 99.5\% of the variance.}  
\label{fig:abnormal_score}
\end{figure} 

Fig.~\ref{fig:abnormal_score} plots the projection residual of both
the training data and the test data, using both the variance-based and
the distance-based subspace methods. In case of the variance-based
approach, we use the number of dimensions required to capture
99.5\% of the variance in the training data as the number of
dimensions for the normal subspace. In case of the distance-based
approach, for each test dataset (comprised of multiple time
windows), we derive a new effective subspace dimension (ESD) to serve
as the dimension of the normal subspace. In
Fig.~\ref{fig:abnormal_score}, the number of dimensions of normal
subspace used for the projections is 11 in the case of the
variance-based approach and a lower number, 8, in the case of
our distance-based approach. As Fig.~\ref{fig:abnormal_score} shows, through qualitative
observation, that both the variance-based and the distance-based
subspace methods are able to distinguish between test data (which
contains anomalous connections) and the training data (which contains
no anomalous connections). 



\begin{figure}[!t]
\centering
\epsfig{file=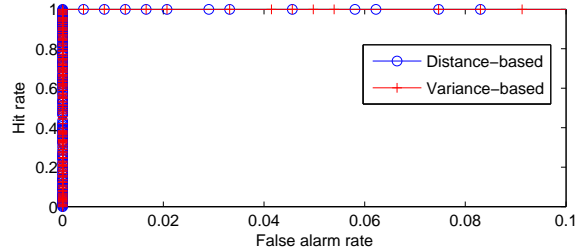,width=3.4in}
\caption{Plots of ROC as a result of the variance-based and distance-based subspace method. Here, the normal subspace dimension in the distance-based subspace method is 8. The normal subspace dimension in the variance-based method is 11 and captures 99.5\% variance.}
\label{fig:RoC1}
\end{figure}

\subsection{Hit rate and false alarm rate}

To evaluate our distance-based algorithm for anomaly detection, we use
the Receiver operating characteristic (ROC) metric to characterize the
number of anomalous datapoints in the Kyoto2006+ dataset that can be
detected using the variance-based and distance-based subspace
method. We compare the projection residual of the Kyoto2006+ dataset
with a threshold, and use the resulting hit rate and the false alarm
rate (or the false positive rate) to make a quantitative assessment of the 
effectiveness of anomaly detection. The hit rate is defined as the
percentage of anomalous connections whose projection residual exceeds
a certain threshold. The false alarm rate is defined as the percentage
of non-anomalous connections whose projection residual exceeds the
threshold. The hit rate and false alarm rate varies when the threshold
changes.

We illustrate the comparative performance of variance-based and
distance-based subspace methods using ROCs in two ways. Note that, as
shown in Fig.~\ref{fig:abnormal_score}, the projection residuals
between the anomalous connections and the normal connections are
distinctive for both of these two subspace methods. As a result, we
can achieve 100\% hit rate with zero false alarm rate when a proper
threshold is used for detection (between 2.33 and 2.70 for the
variance-based method, and between 6.69 and 7.03 for the distance-base
method.) Fig.~\ref{fig:RoC1} shows the ROC comparison between the
variance-based and distance-based subspace methods for various
thresholds in this range. The results show that the distance-based
subspace method achieves a better performance since it yields the same
hit rate and false alarm rates even though it uses a reduced number of
dimensions (8 compared to 11.)

\begin{figure}[!t]
\centering
\epsfig{file=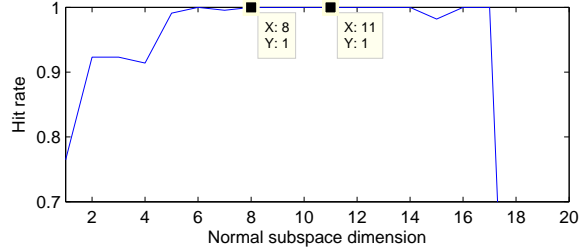,width=3.4in}
\caption{The achieved hit rate (as a percentage) when the false alarm
  rate is fixed at 1\% plotted on the Y-axis against the number of
  dimensions in the normal subspace on the X-axis.}
\label{fig:hit_rate_fa1e-2}
\end{figure}

A second way to compare ROCs is by choosing a threshold such
that the false alarm rate is fixed at a certain percentage, say 1\%,
and then make comparisons between the hit rate under variance-based
and distance-based methods. Fig.~\ref{fig:hit_rate_fa1e-2}
shows the resulting hit rate corresponding to different numbers of
normal subspace dimensions when the false alarm rate is fixed at 1\%. When the normal subspace dimension lies between 8 and 14, the hit rate
is 100\%. However, when the number of dimensions used for the normal
subspace lies outside this range, the performance is worse. In the
variance-based subspace method, the normal subspace dimension 
used for projections and, consequently, the hit rate, can vary
significantly with the choice of the percentage of variance in the
training data that the normal subspace is designed to capture 
(as suggested by Fig.~\ref{fig:hit_rate_fa1e-2}). On the other hand,
the distance-based subspace method depends on the structural changes
in the test data and uses the smallest number of dimensions that
achieves good performance (as also suggested by Fig.~\ref{fig:hit_rate_fa1e-2}).

Thes results in Figs.~\ref{fig:RoC1} and \ref{fig:hit_rate_fa1e-2}
demonstrate a key advantage of the distance-based subspace method over
the variance-based subspace methods. Instead of using a pre-set number
of dimensions based only on the training data, the distance-based
approach uses only the number of dimensions that approximately
maximizes the distinction between the test data and the normal
subspace. For this real-traffic trace, the variance-based approach may
use as many as 14 (in this example, it uses 11 to capture 99.5\% of
the variance), while the distance-based approach uses only 8
dimensions with no appreciable reduction in the hit rate. The
advantage lies in the fact that, unlike the variance-based approach,
the distance-based approach can adapt to the characteristics of the
test data to use only the number of dimensions necessary.

\begin{figure}[!t]
\centering
\epsfig{file=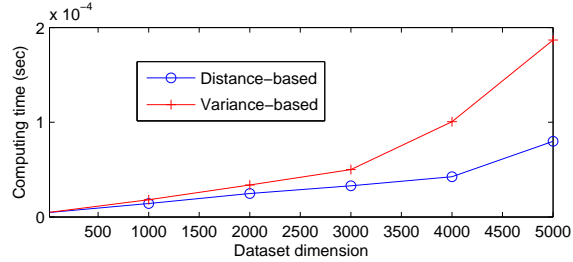,width=3.4in}
\caption{A comparison of the running time for computing the projection residuals using the distance-based and variance-based subspace methods.}
\label{fig:runningTime_calProj}
\end{figure} 

The benefit of using only the number of dimensions necessary is the
ability to achieve a reduced runtime without compromising anomaly
detection. To show this advantage of the distance-based method for
real-time detection, we evaluate the average running time for
computing the projection residual for the distance-based and
variance-based subspace methods. The simulation was implemented in
MATLAB and executed on a server equipped with an AMD Athlon II 3.0 GHz
processor. 
Using synthetic datasets with increasing number of dimensions, we show
in Fig.~\ref{fig:runningTime_calProj} that as the dimension of the
dataset increases, the distance-based method is more efficient in
reducing the overhead of computing the projection residual. More
specifically, the runtime with the distance-based method is less than
half that with variance-based method as the number of features in the
dataset approaches 5000, a very realistic scenario in network management.


\begin{figure}[!t]
\centering
\epsfig{file=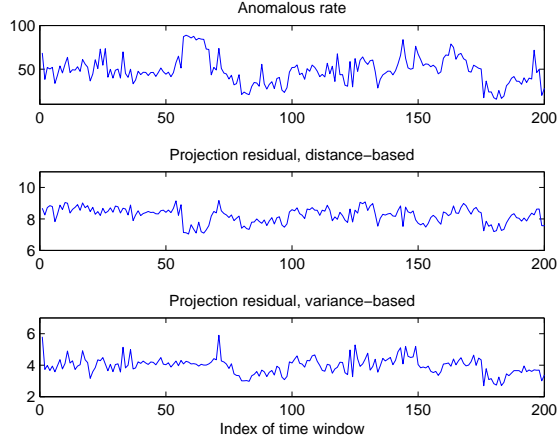,width=3.4in}
\caption{A comparison of the anomalous rate with the projection
  residuals using the distance-based and variance-based subspace
  methods.}  
\label{fig:anomalous_rate_score}
\end{figure} 

\subsection{Information content of residuals}

Besides the real-time adaptability, the distance-based subspace
approach offers a second key advantage over the variance-based
approach --- the projection residual in the
distance-based approach carries more useful information toward
successful anomaly detection. To illustrate this, we will use the 
labels (which identify a connection as part of attack traffic or not)
in the Kyoto dataset to calculate the percentage of anomalous
connections within each time window, and refer to this percentage as
the \textit{anomalous rate}.  

Fig.~\ref{fig:anomalous_rate_score} offers a comparison between the
projection residual computed for each time window in the test data
with the anomalous rate within the same time window. One interesting and
illustrative observation that emerges is that, in general, the anomalous
rate tends to somewhat track the projection residual but not
consistently. A sharp and prolonged increase in the anomalous rate
during time windows 55--65 leads to a reduced projection residual in
the case of distance-based approach, while in the case of
variance-based approach, there is no appreciable change. 

\begin{figure}[!t]
\centering
\epsfig{file=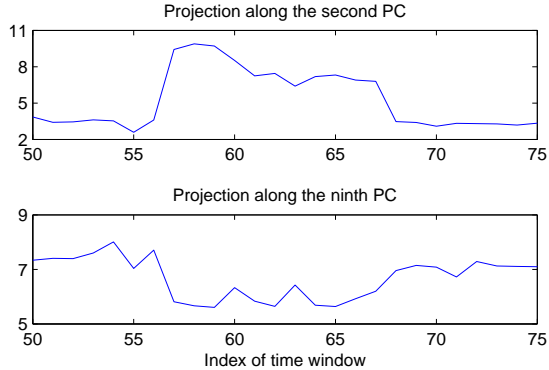,width=3.4in}
\caption{Projection of the test data along the second and the ninth
  principal components of the normal subspace.}  
\label{fig:proj_entropy_v2_v9}
\end{figure}

For a deeper understanding of this phenomenon, we examined each of
the principal components of the test data and the normal subspace in
the time windows 55--65. The second and the ninth principal components
of the training data are pertinent here and projections of the test
data on these principal components are plotted in
Fig.~\ref{fig:proj_entropy_v2_v9}. In time windows with a high anomalous rate, we find that the anomalous patterns are largely along
the second principal component of the normal subspace. However, it
also deviates from the normal subspace in the less significant
principal components, as shown in the figure for the ninth principal
component. 

The variance-based approach, because of its static use of the number
of dimensions based only on training data, does not offer any hint of
the rise in anomalous traffic through the projection residual. The
distance-based approach, on the other hand, shows a sudden change in
the projection residual offering the potential for a richer set of
heuristics for anomaly detection. The distance-based subspace method, by capturing
more information about the test data in its dimensionality reduction,
also transfers some of this information into the pattern of changes
in the projection residual. 

Heuristics for anomaly detection can be further enhanced in the case
of distance-based subspace method because the algorithm, in the
process of computing the effective subspace dimension, also computes
the principal components of the test data. Periodic analysis of these
principal components can offer deeper insights into anomalous
behavior and can, in fact, as shown in Section~\ref{sec:spoofing},
even help detect malicious traffic which tries to mimic normal
traffic in its principal components.  

\subsection{Overcoming normal traffic spoofing}\label{sec:spoofing}

Just like cryptographers are required to assume that attackers have
knowledge of the cryptographic algorithm, system security engineers
have to assume that attackers will have knowledge of the normal
subspace used for anomaly detection. We define a traffic spoofing
attack as one in which an attacker deliberately engineers the profile
of the anomalous traffic so that its most significant principal
components align with those of the normal subspace. The variance-based approach,
because its dimensionality reduction is based only on the normal
subspace, is vulnerable to this type of spoofing attack. The
distance-based approach, however, is able to resist this attack
better; because of its dependence on the principal components of the
test data, an attacker needs to spoof not just the principal components of
the normal subspace but also in the right order, which presents the
attacker a higher hurdle to cross. 

Suppose an anomaly detection system uses the histogram of
packet sizes and protocols as traffic features. In order to launch a
denial-of-service attack that evades detection, the attacker can
design the size of attack packets such that the attack traffic follows
a subset of the principal components in the normal subspace. But,
assume that the order of the principal components of the attack
traffic differ from those of the normal traffic, since preserving the order
would be a more difficult feat for an attacker.

Suppose the normal dataset (BEFORE), without any anomalies, has
$N\times N$ covariance matrix $\Sigma_B = U\Lambda_B U^{\prime}$, where
the principal components are given by $U = V^{\prime} = [\textbf{u}_1,
\dots, \textbf{u}_N]$ with random orthonormal columns, and $\Lambda_B$ is a diagonal
 matrix. Assume
that the variance-based method uses a threshold percentage of captured
variance such that all four of these principal components are in the
normal subspace. Using 99.5\% as the
minimum variance that should be captured in the variance-based
approach, the normal subpace dimension with the variance-based
approach is 8. Assume the injected
 anomalous traffic behaves as the 4-th principal component in the
 normal subspace, and the resulting covariance matrix (AFTER) becomes
 $\Sigma_A = V\Lambda_A V^{\prime}$, where $\Lambda_A$ is also a diagonal
 matrix. Let the principal components of $\Sigma_A$ be given
 by $V = [\textbf{v}_1,\dots,\textbf{v}_N]$, and  
  \[  \textbf{v}_n = \left\{ 
  \begin{array}{l l}
    \textbf{u}_4 & \quad \text{if~}n=3\\
    \textbf{u}_3 & \quad \text{if~}n=4\\
    \textbf{u}_n & \quad \text{otherwise.}
  \end{array} \right.\]
In other words, the injected anomalous traffic changes the order of
importance of these two principal components, $\textbf{u}_3$ and
$\textbf{u}_4$, both of which resides in the normal subspace. 

\begin{figure}[!t]
\centering
\epsfig{file=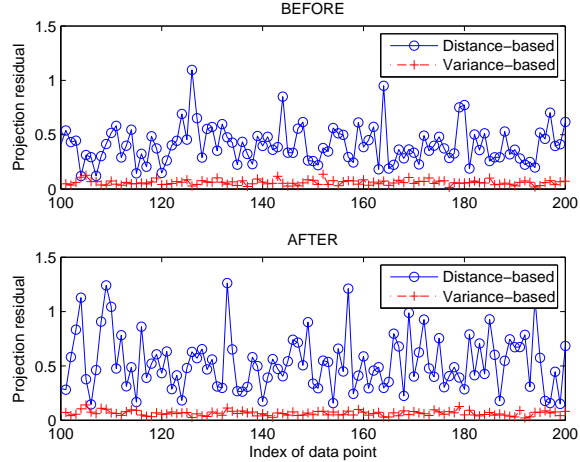,width=3.4in}
\caption{Projection residual of test data without any injected
  anomalies (Top, dataset BEFORE) and with injected anomalies which
  spoof the principal components of the normal subspace (Bottom,
  dataset AFTER)}
\label{fig:projection_3_4}
\end{figure}
 
Upon simulation of this setting, we find that the effective subspace
dimension using the distance-based method is 3. Fig.~\ref{fig:projection_3_4} presents the projection
residual before and after such anomalous traffic is injected. One
thing we observe from Fig.~\ref{fig:projection_3_4} is 
that because the anomalous data resides in the normal subspace, there
is no substantive difference between the projection residual of normal
data and anomalous data (BEFORE and AFTER) when using the variance-based
subspace method. This confirms that the variance-based subspace method
is not able to detect the anomaly that exploits normal traffic spoofing.

On the other hand, the projection residual with the distance-based
subspace method shows a distinct qualitative difference and a larger
value with anomalous traffic. Besides showing that this detection is
possible with only 3 dimensions (while the variance-based approach
uses 8 dimensions without being able to detect it), the distance-based
approach also offers additional information for classification or
mitigation. The \algoname{getESD} algorithm computes, along with the
effective subspace dimension, the most significant principal components of the test
traffic which can be examined for a deeper look into the anomaly
identified by the higher projection residual. In this case, 
one may be able to observe that the order of the principal components
has changed in the test data and that the anomalous traffic may be
attempting to spoof normal traffic.

\subsection{Comparison of centralized and distributed algorithms}
In this subsection, we offer a comparison between the
  performance of the centralized (\algoname{getESD}) and distributed
  (\algoname{getESD-D}) versions of our algorithm. Note that both of
  these versions yield the same mathematical output --- that is, they
  achieve the same accuracy and, therefore, the same level of success
  in anomaly detection. Table~\ref{fig:cost_summary}
  summarizes the comparisons for detecting an anomaly using
  the same workstation as in Fig.~\ref{fig:runningTime_calProj} for our
  simulations. For our simulations, we use synthetic networks modeled
  after realistic campus-wide WiFi backbones, with $N=100$ and
  $N=1000$ and 1Gbps links.


The detection time shown in
  Table~\ref{fig:cost_summary} is the total worst-case runtime for
  computing the projection residual of each data vector for
  detection (in the distributed case, different nodes may converge to
  the answer at different times). As one might expect in a distributed
  algorithm in which detection steps begin with incomplete information
  until convergence is achieved, the worst-case run time of the
  distributed algorithm is longer. However, the distributed algorithm
  comes with several distinct advantages over the centralized version
  as described below.

In the centralized version, the detection occurs at
  the monitoring station. In the distributed version, the detection
  occurs at each node in the network (which is a significant advantage
  in large networks since localized mitigation actions can be applied
  faster and with knowledge of a fuller local context unavailable to a
  distant monitoring station.) The distributed algorithm also removes
  the communication and storage burden at the one monitoring station and
  distributes it across all the nodes. As Table~\ref{fig:cost_summary}
  shows, the communication at the monitoring station in the
  centralized case is significantly higher than the communication at
  each of the nodes in the distributed case --- which is especially
  helpful in the midst of a severe congestion caused by a distributed
  denial-of-service attack.

Storage needs for detection in both the centralized
  and distributed algorithms are primarily that for storing the normal
  subspace data. The only difference is that, in algorithm
  \algoname{getESD-D}, the stored information is evenly distributed among 
  the nodes in the network instead of being concentrated at the
  monitoring station. In general, the monitoring station in
  \algoname{getESD} has to store $N$ times as much data as each node
in \algoname{getESD-D}.

\begin{table}[!t]
\centering
\caption{Cost of algorithms \algoname{getESD} (at the
    monitoring station) and \algoname{getESD-D} (at each node).}
  \vspace{-6pt}
	\begin{tabular}{|c|p{3cm}|r|r|}
	\cline{1-4}
	\multicolumn{1}{ |c| }{\multirow{2}{*}{Metric} } &
	\multicolumn{1}{ c| }{\multirow{2}{*}{$N$} } 
	& \multicolumn{2}{ c| }{Algorithm} \T\B\\ \cline{3-4}
	& &\algoname{getESD} & \algoname{getESD-D} \T\B\\ \cline{1-4}
	\multicolumn{1}{ |c| }{\multirow{2}{*}{\parbox{3cm}{Runtime for\\detection (secs)} }} &
	\multicolumn{1}{ |c| }{100} & 8.36e-5 & 1.22e-4 \T\\ 
	\multicolumn{1}{ |c  }{}                        &
	\multicolumn{1}{ |c| }{1000} & 3.50e-3 & 1.14e-2 \B\\ \cline{1-4}
	\multicolumn{1}{ |c| }{\multirow{2}{*}{\parbox{3cm}{Communication\\cost of detection}} } & 
	\multicolumn{1}{ |c| }{100} & 5.97 KB & 400 B \T\\ 
	\multicolumn{1}{ |c  }{}                        &
	\multicolumn{1}{ |c| }{1000} & 85.64 KB & 560 B \B\\
          \cline{1-4}
	\multicolumn{1}{ |c| }{\multirow{2}{*}{\parbox{3cm}{Storage cost\\of detection}} } & 
	\multicolumn{1}{ |c| }{100} & 78.12 KB & 0.78 KB \T\\ 
	\multicolumn{1}{ |c  }{}                        &
	\multicolumn{1}{ |c| }{1000} & 7.63 MB & 7.81 KB \B\\
          \cline{1-4}
	\end{tabular} 
  \label{fig:cost_summary}
  \vspace{-12pt}
\end{table}

\section{Concluding remarks}\label{sec:conclusion}

Given Big Data trends in both collection and storage, current
state-of-the-art network traffic analysis invariably deals with
high-dimensional datasets of increasingly larger size --- thus, it is
important to derive a low-dimensional structure as a compact
representation of the original dataset. 
In this paper, supported by
theoretical analysis and simulation results using real traffic traces,
we have described a new distance-based approach to dimensionality reduction with
distinct advantages over previously known methods.
These advantages
include (i) improved adaptability to changing patterns in test data so
that we only use the number of dimensions necessary at any given time,
(ii) improved ability to characterize and classify attack traffic
since the dimensionality reduction is based on the observed traffic
and not pre-determined, and (iii) improved resilience against attacks
which try to spoof normal traffic patterns.

Even though we illustrated our distance-based subspace approach
through the end goal of anomaly detection in networks, our
contribution can be employed in any other network management context
where dimensionality reduction of live network traffic is useful. 
The primary
technical contribution of this paper, therefore, is a new general
distance-based method for real-time dimensionality reduction in order
to ease the Big Data challenges of network and system management.

\appendix
\label{sec:appendix}

\subsection{Proof of Theorem \ref{theo:kA-equals-kB}}\label{proof:theorem1}

Without loss of generality, assume $k_A\geq k_B$. The statement of the
theorem is proved if $\theta_{k_A, k_B}(A,B)\leq \theta_{k_B,
  k_B}(A,B)$ and $\theta_{k_A, k_B}(A,B)\leq \theta_{k_A,
  k_A}(A,B)$. The proofs of each of these two cases follows.

{\em Case (i):} Let $\textbf{x}$ denote a column vector of length
$k_B$. Using the definition of matrix norm, we have:
\begin{align*}
\parallel & \, T_{k_A, k_B}(A,B)\parallel^2\\
& = \max_{\forall \textbf{x}, \parallel \textbf{x} \parallel=1} \parallel T_{k_A, k_B}(A,B)\textbf{x} \parallel^2\\
& = \max_{\forall \textbf{x}, \parallel \textbf{x} \parallel=1} \sum_{i=k_A+1}^{N} \parallel\textbf{a}_i^\prime [\textbf{b}_1, \dots, \textbf{b}_{k_B}]\textbf{x}\parallel^2\\
& = \sum_{i=k_A+1}^{N} \parallel\textbf{a}_i^\prime [\textbf{b}_1,
\dots, \textbf{b}_{k_B}]\textbf{x}_{\mathrm{max}}\parallel^2
\end{align*}
where $\textbf{x}_{\mathrm{max}}$ is the column vector with unit norm
that achieves the maximum matrix norm. Similarly, using the definition
of the matrix norm again:
\begin{align*}
\parallel & \, T_{k_B, k_B}(A,B)\parallel^2\\
& = \max_{\forall \textbf{x},\parallel \textbf{x} \parallel=1} \sum_{i=k_B+1}^{N} \parallel\textbf{a}_i^\prime [\textbf{b}_1, \dots, \textbf{b}_{k_B}]\textbf{x}\parallel^2\\
& \geq \sum_{i=k_B+1}^{N} \parallel\textbf{a}_i^\prime [\textbf{b}_1,
\dots, \textbf{b}_{k_B}]\textbf{x}_{\mathrm{max}}\parallel^2\\
& \geq \sum_{i=k_A+1}^{N} \parallel\textbf{a}_i^\prime [\textbf{b}_1,
\dots, \textbf{b}_{k_B}]\textbf{x}_{\mathrm{max}}\parallel^2
=\parallel T_{k_A, k_B}(A,B)\parallel^2.
\end{align*}
Thus we have $\theta_{k_A, k_B}(A,B)\leq \theta_{k_B, k_B}(A,B)$.

{\em Case (ii)}: Let $\textbf{y}$ denote a column vector of length
$k_A$. Using the definition of matrix norm: 
\begin{align*}
\parallel & \, T_{k_A, k_A}(A,B)\parallel^2 = \max_{\forall \textbf{y}, \parallel \textbf{y} \parallel=1} \sum_{i=k_A+1}^{N} \parallel\textbf{a}_i^\prime [\textbf{b}_1, \dots, \textbf{b}_{k_A}]\textbf{y}\parallel^2
\end{align*}
Let $\textbf{z}$ denote a column vector of length $k_A$ with
  $\textbf{x}_{\mathrm{max}}$ followed by
  $k_A-k_B$ zeroes: $\textbf{z}^\prime=[\textbf{x}_{\mathrm{max}}~ 0
  \cdots 0]^\prime$. This vector has unit norm and $[\textbf{b}_1,
  \dots, \textbf{b}_{k_A}]\textbf{z} = [\textbf{b}_1, \dots,
  \textbf{b}_{k_B}]\textbf{x}_{\mathrm{max}}$. Therefore,
\begin{align*}
&\parallel T_{k_A, k_A}(A,B)\parallel^2 ~ \geq
\sum_{i=k_A+1}^{N} \parallel\textbf{a}_i^\prime [\textbf{b}_1, \dots,
\textbf{b}_{k_A}]\textbf{z} \parallel^2 \\
& = \sum_{i=k_A+1}^{N} \parallel\textbf{a}_i^\prime [\textbf{b}_1,
\dots, \textbf{b}_{k_B}]\textbf{x}_{\mathrm{max}}\parallel^2  = \, \parallel T_{k_A, k_B}(A,B)\parallel^2.
\end{align*}
Thus we have $\theta_{k_A, k_B}(A,B)\leq \theta_{k_A, k_A}(A,B)$.

\hfill{\small $\blacksquare$}

\subsection{Proof of Theorem \ref{theo:smallest-singular-value}}\label{proof:theorem2}

First, we use $P_{i,j}$ to express $T_{k,k}(A,B)$ and $T^{\prime}_{k,k}(A,B)T_{k,k}(A,B)$.
\begin{align*}
T_{k,k}(A,B) &= (I-\sum_{i=1}^{k} \textbf{a}_i\times\textbf{a}_i^\prime)[\textbf{b}_1, \dots, \textbf{b}_{k}] \\
&= [\textbf{b}_1, \dots, \textbf{b}_{k}] - \sum_{i=1}^{k} \textbf{a}_i\times[\textbf{a}_i^\prime\textbf{b}_1, \dots, \textbf{a}_i^\prime\textbf{b}_{k}] \\
&= [\textbf{b}_1, \dots, \textbf{b}_{k}] - \sum_{i=1}^{k} \textbf{a}_i[P_{i,1}, \dots, P_{i,k}]
\end{align*}
Similarly,
\begin{align*}
T^{\prime}_{k,k}(A,B)T_{k,k}(A,B)  = I_{k\times k}-P_{1:k,1:k}^{\prime} P_{1:k,1:k}
\end{align*}
A more detailed proof with all intermediate steps is provided in
\cite{Huang2015-thesis}.

The value of $\sin(\theta_{k, k}(A,B))$, defined earlier as
$\parallel\,T_{k,k}(A,B)\,\parallel$, is the first singular value of
$T_{k,k}(A,B)$ and also the square root of the largest eigenvalue of
$T^{\prime}_{k,k}(A,B)T_{k,k}(A,B)$. Following the previous analysis,
therefore, we have:
\begin{align*}
\sin^2(\theta_{k, k}(A,B)) &= \,\parallel T_{k,k}(A,B)\parallel^2\\
&= \sigma_1^2(T_{k,k}(A,B))\\
&= \lambda_1(T^{\prime}_{k,k}(A,B)T_{k,k}(A,B))\\
&= \lambda_1(I_{k\times k}-P_{1:k,1:k}^{\prime} P_{1:k,1:k})\\
&= 1 - \lambda_k(P_{1:k,1:k}^{\prime} P_{1:k,1:k})\\
&= 1 - \sigma_k^2(P_{1:k,1:k})
\end{align*} 
where $\lambda_i$ stands for the $i$-th eigenvalue and $\sigma_i$ for the $i$-th singular value.
As a result, the following holds:
\begin{align*}
\cos(\theta_{k, k}(A,B)) = \sigma_k(P_{1:k,1:k})
\end{align*}

\hfill{\small $\blacksquare$}

\subsection{Proof of Theorem \ref{theo:non-decreasing-to-1}}\label{proof:theorem3}

Consider an $N\times N$ matrix $P$ built as in
Eq.~(\ref{eq:projection}). Let $\textbf{u}_i$ denote an $N\times 1$
vector with $i$-th entry being non-zero:
  \[  \textbf{u}_i(k) = \left\{ 
  \begin{array}{l l}
    1 & \quad \text{if~}k=i\\
    0 & \quad \text{if~}k\neq i
  \end{array} \right.\]

Now, letting $\textbf{p}_{i}$ denote the $i$-th column of $P$, and
letting $I_{N\times N}$ denote an identity matrix of size $N\times N$,
we have:
\begin{align*}
\textbf{u}_i^{\prime}P^{\prime} P\textbf{u}_j & = \textbf{p}_{i}^{\prime}\textbf{p}_{j}\\
& = \sum_{k=1}^{N}P_{k,i}P_{k,j} =
  \sum_{k=1}^{N}\textbf{b}_i^{\prime}\textbf{a}_k\textbf{a}_k^{\prime}\textbf{b}_j\\
& =
  \textbf{b}_i^{\prime}(\sum_{k=1}^{N}\textbf{a}_k\textbf{a}_k^{\prime})\textbf{b}_j
  = \textbf{b}_i^{\prime}I_{N\times N}\textbf{b}_j = \textbf{b}_i^{\prime}\textbf{b}_j
\end{align*}
Therefore, we have
\begin{equation}  \textbf{u}_i^{\prime}P^{\prime} P\textbf{u}_j = \left\{ 
  \begin{array}{l l}
    1 & \quad \text{if~}i=j\\
    0 & \quad \text{if~}i\neq j
  \end{array} \right.
\label{lemma1}
\end{equation}

Now, for simplicity, denote $P_{1:k,1:k}$ as $P_k$. We prove the
result in Theorem~\ref{theo:non-decreasing-to-1} by first proving that
$\sigma_1(P_{k})>\sigma_1(P_{k-1})$ and then proving that
$\sigma_1(P_{k})\leq 1$. 

To prove that $\sigma_1(P_{k})>\sigma_1(P_{k-1})$, note that:
\begin{equation} \label{eq:Pk}
    P_{k} = \begin{bmatrix}
    	P_{k-1} & \textbf{u} \\
    	\textbf{v}^{\prime} & c
     \end{bmatrix}
\end{equation}
where $\textbf{u}^{\prime} = [\textbf{a}^{\prime}_1\textbf{b}_k, \dots, \textbf{a}^{\prime}_{k-1}\textbf{b}_k]$, 
$\textbf{v}^{\prime} = [\textbf{a}^{\prime}_k\textbf{b}_1, \dots, \textbf{a}^{\prime}_{k}\textbf{b}_{k-1}]$, and $c = \textbf{a}^{\prime}_{k}\textbf{b}_{k}$.
Then, $P^{\prime}_{k}P_{k}$ can be constructed as follows:
\begin{align}\label{eq:PkPk}
    &P^{\prime}_{k}P_{k} 
    = 
    \begin{bmatrix}
    	P^{\prime}_{k-1} & \textbf{v} \\
		\textbf{u}^{\prime} & c
    \end{bmatrix}
    \begin{bmatrix}
		P_{k-1} & \textbf{u} \\
		\textbf{v}^{\prime} & c
    \end{bmatrix}\nonumber\\
    =& 
    \begin{bmatrix}
    	P^{\prime}_{k-1}P_{k-1}+\textbf{v}\textbf{v}^{\prime} & P^{\prime}_{k-1}\textbf{u}+c\textbf{v} \\
		\textbf{u}^{\prime}P_{k-1}+c\textbf{v}^{\prime} & \textbf{u}^{\prime}\textbf{u}+c^2
    \end{bmatrix}
\end{align}

Since $\sigma_1(P_{k})$ is the first singular value of $P_{k}$, it
implies the following: first, $\sigma^2_1(P_{k})$ is an eigenvalue of
$P^{\prime}_{k}P_{k}$, i.e., $P^{\prime}_{k}P_{k}\textbf{z}_{k} =
\sigma^2_1(P_{k})\textbf{z}_{k}$, where $\textbf{z}_{k}$ is the
corresponding eigenvector with unit norm; second, $\parallel
P_{k}\textbf{z} \parallel \, \leq \sigma_1(P_{k})$ 
for any vector $\textbf{z}$ with unit norm. Thus, we have:
\begin{align*}
\sigma_1^2(P_{k}) & \geq  \begin{bmatrix}\textbf{z}^{\prime}_{k-1} & 0
    \end{bmatrix}
 P^{\prime}_{k}P_{k}
  \begin{bmatrix}
 	\textbf{z}_{k-1}\\
 	0
 \end{bmatrix}  \\ 
& = \begin{bmatrix}
  	\textbf{z}^{\prime}_{k-1} & 0
 \end{bmatrix}
  \begin{bmatrix}
 	P^{\prime}_{k-1}P_{k-1}\textbf{z}_{k-1}+\textbf{v}\textbf{v}^{\prime}\textbf{z}_{k-1}\\
 	\textbf{u}^{\prime}P_{k-1}\textbf{z}_{k-1}+c\textbf{v}^{\prime}\textbf{z}_{k-1}
 \end{bmatrix}  \\
& = \textbf{z}^{\prime}_{k-1}P^{\prime}_{k-1}P_{k-1}\textbf{z}_{k-1}+\textbf{z}^{\prime}_{k-1}\textbf{v}\textbf{v}^{\prime}\textbf{z}_{k-1}\\
& = \textbf{z}^{\prime}_{k-1}\sigma_1^2(P_{k-1})\textbf{z}_{k-1}+(\textbf{v}^{\prime}\textbf{z}_{k-1})^2\\
& \geq \sigma_1^2(P_{k-1})\textbf{z}^{\prime}_{k-1}\textbf{z}_{k-1}\\
& = \sigma_1^2(P_{k-1}) 
\end{align*}
Since $\sigma_1(P_{k})$ is non-negative, we have $\sigma_1(P_{k}) \geq \sigma_1(P_{k-1})$.

Now, to prove that $\sigma_1(P_{k})\leq1$, note that for any
$\textbf{x}\in R^{N\times 1}$ with unit norm, we can represent it as a
linear combination of $\textbf{u}_i$, $\textbf{x} = \sum_{k=1}^{N}
x_i\textbf{u}_i$, with $\sum_{k=1}^N x_i^2 = 1$. Using the result from
Equation~(\ref{lemma1}), we have 
\begin{align*}
\parallel P\textbf{x}\parallel^2 & = \textbf{x}^{\prime}P^{\prime} P\textbf{x} \\
& = \left(\sum_{i=1}^{N} x_i\textbf{u}_i^{\prime}\right) P^{\prime}P \sum_{j=1}^{N} x_j\textbf{u}_j\\
& = \sum_{i=1}^{N}\left(\sum_{j=1}^{N} x_i x_j \textbf{u}_i^{\prime} P^{\prime}P \textbf{u}_j\right)
= \sum_{i=1}^{N} \left(x_i^2\right) = 1
\end{align*}
Therefore, $\sigma_1(P)=\dots =\sigma_N(P)$ and $\sigma_1(P) = \max \parallel P\textbf{x}\parallel = 1$. As a result,
\begin{align*}
&\sigma_1(P_{k})\leq \sigma_1(P_{N})= \sigma_1(P)= 1
\end{align*}
\hfill{\small $\blacksquare$}

\subsection{Proof of Theorem \ref{theo:getESD-complexity}}\label{proof:theorem4}
The proof of Theorem \ref{theo:non-decreasing-to-1} above demonstrates that one can construct a
projection matrix in search of the optimal subspace dimension. First,
based on Equations (\ref{eq:Pk}) and (\ref{eq:PkPk}), let's look at
the computational complexity of constructing $P_{1:k,1:k}$ and
$P^{\prime}_{1:k,1:k}P_{1:k,1:k}$ with the knowledge of 
$P^{\prime}_{1:k-1,1:k-1}P_{1:k-1,1:k-1}$ and $\textbf{a}_1, \dots,
\textbf{a}_k, \textbf{b}_1, \dots, \textbf{b}_k$.  

The complexity of calculating $\textbf{u}$, $\textbf{v}$
and $c$ is $\mathcal{O}(kN)$. The complexity of constructing
$P^{\prime}_{1:k,1:k}P_{1:k,1:k}$ is $\mathcal{O}(k^2)$.

Now, the complexity of calculating the eigenvalue of 
$P^{\prime}_{1:k,1:k}P_{1:k,1:k}$ using power iteration method is
$\mathcal{O}(Zk^2)$ with $Z$ being the number of iterations necessary
for convergence. Since $k<N$, the overall complexity of algorithm
\algoname{getESD} in $k$ loops is $\mathcal{O}(Zk^3+k^2N)$. 

\hfill{\small $\blacksquare$}

%
%
%
%

\subsection{Proof of Theorem \ref{theo:getESD-D-comp-complexity}}\label{proof:theorem8}

The most computationally expensive part is in calculating $\textbf{z}_n^{(k)}$. When the size of measurements is $M$, the length of $\textbf{x}^{\prime}_n \textbf{v}_n^{(k)}$ is also $M$. Assuming the maximum degree of each node is $\Delta$, then the process of average consensus in $S$ steps has complexity of $\mathcal{O}(\Delta S)$ for each entry of $\textbf{x}^{\prime}_n \textbf{v}_n^{(k)}$. Overall, the computational complexity of the distributed power iteration method running for $p$ steps is $\mathcal{O}(pM\Delta S)$. As a result, the overall computational complexity of \algoname{getESD-D} using $\mathcal{O}(k)$ power iterations is $\mathcal{O}(kpM\Delta S)$.

\hfill{\small $\blacksquare$}

\subsection{Proof of Theorem \ref{theo:getESD-D-comm-complexity}}
\label{proof:theorem9}

We consider the number of messages that one node needs to send when the algorithm is running. Again, let the maximum degree of each node be $\Delta$. At the very step, the node needs to send its observation $\textbf{x}_n$ to its neighbors, leading to communication cost of $\mathcal{O}(M\Delta)$. After that, this node only needs to update its neighbors about its estimate $\textbf{v}_n^{(k)}$. The number of messages sent for each power iteration is $\mathcal{O}(\Delta S)$. Overall, the communication complexity of the distributed power iteration method running for $p$ steps is $\mathcal{O}(M\Delta + p\Delta S)$. The overall communication complexity of \algoname{getESD-D} using $\mathcal{O}(k)$ power iterations is $\mathcal{O}(kM\Delta + kp\Delta S)$.

\hfill{\small $\blacksquare$}

\bibliographystyle{IEEEtran}
\bibliography{HuangSethuKandasamy}

\begin{thebibliography}{10}
\providecommand{\url}[1]{#1}
\csname url@samestyle\endcsname
\providecommand{\newblock}{\relax}
\providecommand{\bibinfo}[2]{#2}
\providecommand{\BIBentrySTDinterwordspacing}{\spaceskip=0pt\relax}
\providecommand{\BIBentryALTinterwordstretchfactor}{4}
\providecommand{\BIBentryALTinterwordspacing}{\spaceskip=\fontdimen2\font plus
\BIBentryALTinterwordstretchfactor\fontdimen3\font minus
  \fontdimen4\font\relax}
\providecommand{\BIBforeignlanguage}[2]{{%
\expandafter\ifx\csname l@#1\endcsname\relax
\typeout{** WARNING: IEEEtran.bst: No hyphenation pattern has been}%
\typeout{** loaded for the language `#1'. Using the pattern for}%
\typeout{** the default language instead.}%
\else
\language=\csname l@#1\endcsname
\fi
#2}}
\providecommand{\BIBdecl}{\relax}
\BIBdecl

\bibitem{BhuBha2014}
M.~Bhuyan, D.~Bhattacharyya, and J.~Kalita, ``Network anomaly detection:
  Methods, systems and tools,'' \emph{IEEE Commun. Surveys Tuts.}, vol.~16,
  no.~1, pp. 303--336, 2014.

\bibitem{YenOpr2013}
T.-F. Yen, A.~Oprea, K.~Onarlioglu, T.~Leetham, W.~Robertson, A.~Juels, and
  E.~Kirda, ``Beehive: Large-scale log analysis for detecting suspicious
  activity in enterprise networks,'' in \emph{ACM Ann. Comput. Security Appl.
  Conf.}, 2013, pp. 199--208.

\bibitem{Jolliffe2002}
I.~T. Jolliffe, \emph{Principal Component Analysis}, 2nd~ed.\hskip 1em plus
  0.5em minus 0.4em\relax New York: Springer, 2002.

\bibitem{Lakhina2004}
A.~Lakhina, M.~Crovella, and C.~Diot, ``Diagnosing network-wide traffic
  anomalies,'' \emph{SIGCOMM Comput. Commun. Rev.}, vol.~34, no.~4, pp.
  219--230, Aug. 2004.

\bibitem{Ringberg2007}
H.~Ringberg, A.~Soule, J.~Rexford, and C.~Diot, ``Sensitivity of {PCA} for
  traffic anomaly detection,'' \emph{SIGMETRICS Perform. Eval. Rev.}, vol.~35,
  no.~1, pp. 109--120, Jun. 2007.

\bibitem{Yeung2007}
D.~Yeung, S.~Jin, and X.~Wang, ``Covariance-matrix modeling and detecting
  various flooding attacks,'' \emph{IEEE Trans. Syst., Man, Cybern. A, Syst.,
  Humans}, vol.~37, no.~2, pp. 157--169, 2007.

\bibitem{vlassis2005gossip}
N.~Vlassis, Y.~Sfakianakis, and W.~Kowalczyk, ``Gossip-based greedy gaussian
  mixture learning,'' in \emph{Advances in Informatics}.\hskip 1em plus 0.5em
  minus 0.4em\relax Springer, 2005, pp. 349--359.

\bibitem{Mandjes2005}
M.~Mandjes, I.~Saniee, and A.~L. Stolyar, ``Load characterization and anomaly
  detection for voice over {IP} traffic,'' \emph{IEEE Trans. Neural Netw.},
  vol.~16, no.~5, pp. 1019--1026, 2005.

\bibitem{Freire2008}
E.~Freire, A.~Ziviani, and R.~Salles, ``Detecting {VoIP} calls hidden in web
  traffic,'' \emph{IEEE Trans. Netw. Service Manag.}, vol.~5, no.~4, pp.
  204--214, December 2008.

\bibitem{Thing2009}
V.~Thing, M.~Sloman, and N.~Dulay, ``Locating network domain entry and exit
  point/path for {DDoS} attack traffic,'' \emph{IEEE Trans. Netw. Service
  Manag.}, vol.~6, no.~3, pp. 163--174, September 2009.

\bibitem{Xie2009}
Y.~Xie and S.~zheng Yu, ``A large-scale hidden semi-{M}arkov model for anomaly
  detection on user browsing behaviors,'' \emph{IEEE/ACM Trans. Netw.},
  vol.~17, no.~1, pp. 54--65, Feb 2009.

\bibitem{Paschalidis2009}
I.~Paschalidis and G.~Smaragdakis, ``Spatio-temporal network anomaly detection
  by assessing deviations of empirical measures,'' \emph{IEEE/ACM Trans.
  Netw.}, vol.~17, no.~3, pp. 685--697, June 2009.

\bibitem{Thottan2003}
M.~Thottan and C.~Ji, ``Anomaly detection in {IP} networks,'' \emph{IEEE Trans.
  Signal Process.}, vol.~51, no.~8, pp. 2191--2204, 2003.

\bibitem{Lakhina2005}
A.~Lakhina, M.~Crovella, and C.~Diot, ``Mining anomalies using traffic feature
  distributions,'' \emph{SIGCOMM Comput. Commun. Rev.}, vol.~35, no.~4, pp.
  217--228, Aug. 2005.

\bibitem{Tavallaee2008}
M.~Tavallaee, W.~Lu, S.~A. Iqbal, A.~Ghorbani \emph{et~al.}, ``A novel
  covariance matrix based approach for detecting network anomalies,'' in
  \emph{IEEE Commun. Netw. Services Res. Conf.}, 2008, pp. 75--81.

\bibitem{Kind2009}
A.~Kind, M.~P. Stoecklin, and X.~Dimitropoulos, ``Histogram-based traffic
  anomaly detection,'' \emph{IEEE Trans. Netw. Service Manag.}, vol.~6, no.~2,
  pp. 110--121, 2009.

\bibitem{DAlconzo2009}
A.~D. Alconzo, A.~Coluccia, F.~Ricciato, and P.~Romirer-Maierhofer, ``A
  distribution-based approach to anomaly detection and application to {3G}
  mobile traffic,'' in \emph{IEEE Glob. Telecom. Conf.}, 2009, pp. 1--8.

\bibitem{Callegari2011}
C.~Callegari, L.~Gazzarrini, S.~Giordano, M.~Pagano, and T.~Pepe, ``A novel
  {PCA}-based network anomaly detection,'' in \emph{IEEE Intl. Conf. Commun.},
  2011, pp. 1--5.

\bibitem{Nyalkalkar2011}
K.~Nyalkalkar, S.~Sinha, M.~Bailey, and F.~Jahanian, ``A comparative study of
  two network-based anomaly detection methods,'' in \emph{IEEE Intl. Conf.
  Comput. Commun. (INFOCOM)}, 2011, pp. 176--180.

\bibitem{Pascoal2012}
C.~Pascoal, M.~R. de~Oliveira, R.~Valadas, P.~Filzmoser, P.~Salvador, and
  A.~Pacheco, ``Robust feature selection and robust {PCA} for {I}nternet
  traffic anomaly detection,'' in \emph{IEEE Intl. Conf. Comput. Commun.
  (INFOCOM)}, 2012, pp. 1755--1763.

\bibitem{Mateos2012}
G.~Mateos and G.~B. Giannakis, ``Robust {PCA} as bilinear decomposition with
  outlier-sparsity regularization,'' \emph{IEEE Trans. Sig. Process.}, vol.~60,
  no.~10, pp. 5176--5190, 2012.

\bibitem{Kudo2013}
T.~Kudo, T.~Morita, T.~Matsuda, and T.~Takine, ``{PCA}-based robust anomaly
  detection using periodic traffic behavior,'' in \emph{IEEE Intl. Conf.
  Commun. Wksp. (ICC)}, 2013, pp. 1330--1334.

\bibitem{Bur2010}
C.~J.~C. Burges, \emph{Dimension reduction: A guided tour}.\hskip 1em plus
  0.5em minus 0.4em\relax Now Publishers, Inc., 2010.

\bibitem{Jelasity2007}
M.~Jelasity, G.~Canright, and K.~Eng{\o}-Monsen, ``Asynchronous distributed
  power iteration with gossip-based normalization,'' in \emph{Euro-Par 2007
  Parallel Process.}\hskip 1em plus 0.5em minus 0.4em\relax Springer, 2007, pp.
  514--525.

\bibitem{Bertrand2012}
A.~Bertrand and M.~Moonen, ``Power iteration-based distributed total least
  squares estimation in ad hoc sensor networks,'' in \emph{IEEE Intl. Conf.
  Acoust., Speech, Signal Process. (ICASSP)}, 2012, pp. 2669--2672.

\bibitem{Meng2012}
Z.~Meng, A.~Wiesel, and A.~O. Hero~III, ``Distributed principal component
  analysis on networks via directed graphical models,'' in \emph{IEEE Intl.
  Conf. Acoust., Speech, Signal Process. (ICASSP)}, 2012, pp. 2877--2880.

\bibitem{Huang2007}
L.~Huang, X.~Nguyen, M.~Garofalakis, M.~I. Jordan, A.~Joseph, and N.~Taft,
  ``In-network {PCA} and anomaly detection,'' in \emph{Adv. Neural Inf.
  Process. Sys.}, 2006, pp. 617--624.

\bibitem{Wiesel2009}
A.~Wiesel and A.~O. Hero, ``Decomposable principal component analysis,''
  \emph{IEEE Trans. Signal Process.}, vol.~57, no.~11, pp. 4369--4377, 2009.

\bibitem{Boyd2006}
S.~Boyd, A.~Ghosh, B.~Prabhakar, and D.~Shah, ``Randomized gossip algorithms,''
  \emph{IEEE/ACM Trans. Netw. (TON)}, vol.~14, no.~SI, pp. 2508--2530, 2006.

\bibitem{Dimakis2010}
A.~G. Dimakis, S.~Kar, J.~M. Moura, M.~G. Rabbat, and A.~Scaglione, ``Gossip
  algorithms for distributed signal processing,'' \emph{Proc. IEEE}, vol.~98,
  no.~11, pp. 1847--1864, 2010.

\bibitem{Li2011}
L.~Li, A.~Scaglione, and J.~H. Manton, ``Distributed principal subspace
  estimation in wireless sensor networks,'' \emph{IEEE J. Sel. Topics Signal
  Process.}, vol.~5, no.~4, pp. 725--738, 2011.

\bibitem{bjorck1973numerical}
A.~Bj{\"o}rck and G.~H. Golub, ``Numerical methods for computing angles between
  linear subspaces,'' \emph{Math. comput.}, vol.~27, no. 123, pp. 579--594,
  1973.

\bibitem{golub2012matrix}
G.~H. Golub and C.~F. Van~Loan, \emph{Matrix computations}, 4th~ed.\hskip 1em
  plus 0.5em minus 0.4em\relax JHU Press, 2012.

\bibitem{Huang2015}
T.~Huang, H.~Sethu, and N.~Kandasamy, ``A fast algorithm for detecting
  anomalous changes in network traffic,'' in \emph{Intl. Conf. Netw. Service
  Manag.(CNSM)}, 2015.

\bibitem{CAIDA}
\BIBentryALTinterwordspacing
{CAIDA}, ``The {CAIDA} {A}nonymized {I}nternet {T}races 2013.'' [Online].
  Available:
  \url{{http://www.caida.org/data/passive/passive\_2013\_dataset.xml}}
\BIBentrySTDinterwordspacing

\bibitem{KyotoData}
\BIBentryALTinterwordspacing
{Kyoto University}, ``{Traffic Data from Kyoto University's Honeypots}.''
  [Online]. Available: \url{{http://www.takakura.com/Kyoto\_data/}}
\BIBentrySTDinterwordspacing

\bibitem{Huang2015-thesis}
T.~Huang, ``Adaptive sampling and statistical inference for anomaly
  detection,'' Ph.D. dissertation, Drexel University, 2015.

\end{thebibliography}
\end{document}